\documentclass[runningheads]{llncs}
\usepackage{graphicx}
\usepackage{comment}
\usepackage{amsmath,amssymb} 
\usepackage{color}
\renewcommand{\vec}[1]{\boldsymbol{#1}}
\newcommand{\mat}[1]{\mathbf{#1}}
\newcommand{\set}[1]{\mathcal{#1}}

\newcommand{\template}[0]{\mat{T}}

\newcommand{\blendweights}[0]{\mat{W}}

\newcommand{\pose}[0]{\vec{\theta}}
\newcommand{\shape}[0]{\vec{\beta}}
\newcommand{\size}[0]{\vec{\delta}}

\newcommand{\offsets}[0]{\mathbf{D}}

\newcommand{\smpl}[0]{M}
\newcommand{\posefun}[0]{T}
\newcommand{\blendfun}[0]{W}
\newcommand{\offsetfun}[0]{B}
\newcommand{\jointfun}[0]{J}

\newcommand{\posenet}[0]{f_w^{\pose}}
\newcommand{\shapenet}[0]{f_w^{\shape}}
\newcommand{\garnet}[0]{f_w}

\newcommand{\loss}[0]{\mathcal{L}}

\newcommand{\netweight}[0]{w}

\usepackage{bbold}
\usepackage{subfig}
\usepackage{hyperref}
\usepackage{float}
\hypersetup{
    colorlinks=true,
    linkcolor=blue,
    filecolor=magenta,      
    urlcolor=magenta,
}

\newcommand{\GPM}[1]{{\textcolor{red}{[\textbf{GPM:} #1]}}}
\newcommand{\GT}[1]{{\textcolor{blue}{[\textbf{GT:} #1]}}}
\newcommand{\TT}[1]{{\textcolor{cyan}{[\textbf{TT:} #1]}}}
\newcommand{\BB}[1]{{\textcolor{magenta}{[\textbf{BB:} #1]}}}
\renewcommand{\GPM}[1]{}
\renewcommand{\GT}[1]{}
\renewcommand{\TT}[1]{}
\renewcommand{\BB}[1]{}

\begin{document}
\pagestyle{headings}
\mainmatter
\def\ECCVSubNumber{4732}  

\title{SIZER: A Dataset and Model for Parsing 3D Clothing and Learning Size Sensitive 3D Clothing} 

\titlerunning{SIZER}
%
\author{Garvita Tiwari\inst{1}\and
Bharat Lal Bhatnagar\inst{1}\and
Tony Tung\inst{2} \and
Gerard Pons-Moll\inst{1}}

\authorrunning{Tiwari et al.}
%
\institute{MPI for Informatics, Saarland Informatics Campus, Germany \and
Facebook Reality Labs, Sausalito, USA
\email{\{gtiwari,bbhatnag,gpons\}@mpi-inf.mpg.de, tony.tung@fb.com}}

\maketitle

\begin{abstract}



\begin{figure}[h]
	\centering
	\includegraphics[width=0.99\textwidth]{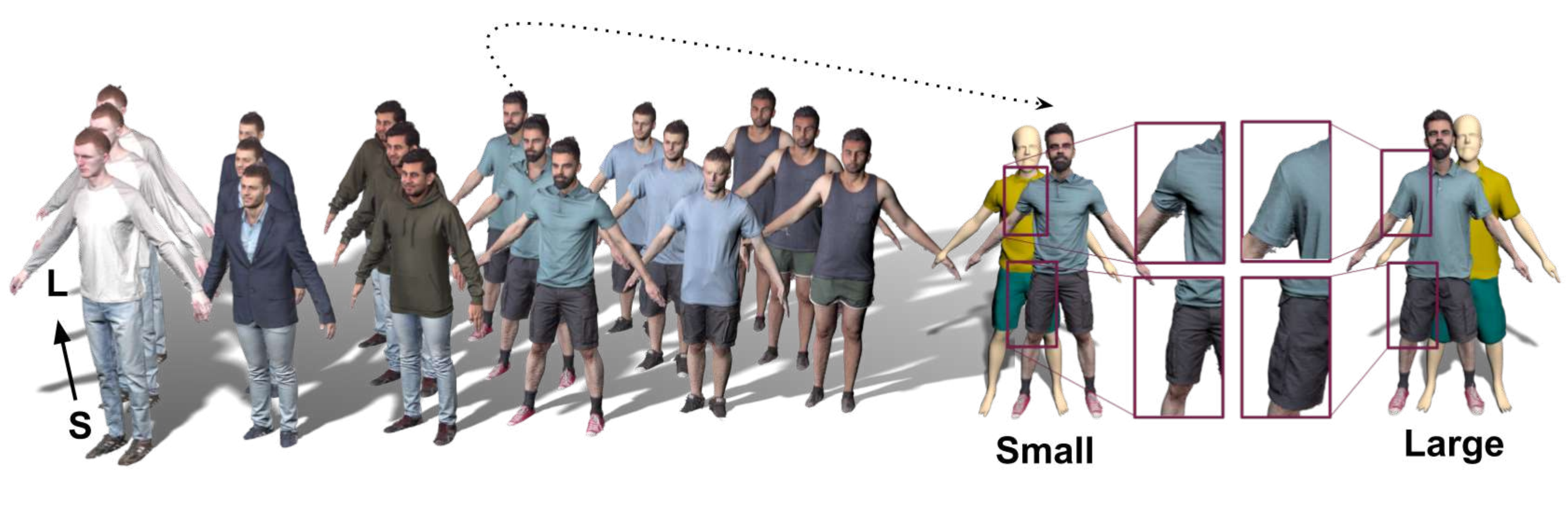}
	\caption{\emph{SIZER} dataset of people with clothing size variation. (\emph{Left}): 3D Scans of people captured in different clothing styles and \emph{sizes}. (\emph{Right}): T-shirt and short pants for sizes small and large, which are registered to a common template.}
	\label{fig:teaser}
\end{figure}
While models of 3D clothing learned from real data exist, no method can predict clothing deformation as a function of garment size.
In this paper, we introduce SizerNet to predict 3D clothing conditioned on human body shape and garment size parameters, and ParserNet to infer garment meshes and shape under clothing with personal details in a single pass from an input mesh.
SizerNet allows to estimate and visualize the dressing effect of a garment in various sizes, and ParserNet allows to edit clothing of an input mesh directly, removing the need for scan segmentation, which is a challenging problem in itself. To learn these models, we introduce the \emph{SIZER} dataset of clothing size variation which includes $100$ different subjects wearing casual clothing items in various sizes, totaling to approximately $2000$ scans. This dataset includes the scans, registrations to the SMPL model, scans segmented in clothing parts, garment category and size labels.
Our experiments show better parsing accuracy and size prediction than baseline methods trained on \emph{SIZER}.
The code, model and dataset will be released for research purposes at: \url{https://virtualhumans.mpi-inf.mpg.de/sizer/}. \GPM{URL or reference to URL on references}

\end{abstract}

\section{Introduction}

Modeling how 3D clothing fits on the human body as a function of size has numerous applications in 3D content generation (e.g., AR/VR, movie, video games, sport), clothing size recommendation (e.g., e-commerce), computer vision for fashion, and virtual try-on. It is estimated that retailers lose up to $\$600$ billion each year due to sales returns as it is currently difficult to purchase clothing online without knowing how it will fit~\cite{ihl,return_article}. 

Predicting how clothing fits as a function of body shape and garment size is an extremely challenging task. Clothing interacts with the body in complex ways, and fit is a non-linear function of size and body shape. Furthermore, \emph{clothing fit differences with size are subtle}, but they can make a difference when purchasing clothing online.   
Physics based simulation is still the most commonly used technique because it generalizes well, but unfortunately, it is difficult to adjust its parameters to achieve a realistic result, and it can be computationally expensive.

While there exist several works that learn how clothing deforms as a function of pose~\cite{laehner2018deepwrinkles}, or pose and shape~\cite{laehner2018deepwrinkles,santesteban2019virtualtryon,gundogdu19garnet,patel2020,ma20autoenclother},
there are few works modeling how garments drape as a function of size. Recent works learn a space of styles~\cite{garmentdesign_Wang_SA18,patel2020} from physics simulations, but their aim
is plausibility, and therefore they can not predict how a \emph{real garment} will deform on a real body. 

What is lacking is (1) a 3D dataset of people wearing the same garments in different sizes and (2) a data-driven model \emph{learned from real scans} which varies with sizing and body shape. 
In this paper, we introduce the \emph{SIZER} dataset, the first dataset of scans of people in different garment sizes featuring approximately $2000$ scans, $100$ subjects and $10$ garments worn by subjects in four different sizes.
Using the \emph{SIZER} dataset we learned a Neural Network model, which we refer to as \emph{SizerNet}, which given a body shape and a garment, can predict how the garment drapes on the body as a function of size. Learning \emph{SizerNet} requires to map scans to a registered \emph{multi-layer meshes} -- separate meshes for body shape, and top and bottom garments.
This requires segmenting the 3D scans, and estimating their body shape under clothing, and registering the garments across the dataset, which we obtain using the method explained in~\cite{bhatnagar2019mgn,ponsmoll2017clothcap}. 
From the multi-layer meshes, we learn an encoder to map the input mesh to a latent code, and a decoder which additionally takes the body shape parameters of SMPL~\cite{SMPL:2015}, the size label (S, M, L, XL) of the input garment, and the desired size of the output, to predict the output garment as a displacement field to a template.

Although visualizing how an existing garment fits on a body as a function of size is already useful for virtual try-on applications, we would also like to change the size of garments in existing 3D scans. Scans however, are just pointclouds, and parsing them into a multi-layer representation at test time using~\cite{bhatnagar2019mgn,ponsmoll2017clothcap} requires segmentation, which sometimes requires manual intervention. Therefore, we propose \emph{ParserNet}, which automatically maps a single mesh registration (SMPL deformed to the scan) to multi-layer meshes with a single feed-forward pass. \emph{ParserNet}, not only segments the single mesh registration, but it reparameterizes the surface so that it is coherent with common garment templates.
The output multi-layer representation of \emph{ParserNet} is powerful as it allows simulation and editing meshes separately. 
Additionally, the tandem of \emph{SizerNet} and \emph{ParserNet} allows us to edit the size of clothing directly on the mesh, allowing shape manipulation applications never explored before.


In summary, our contributions are:
\begin{itemize}
	\item [$\bullet$] \emph{SIZER} dataset: A dataset of clothing size variation of approximately $2000$ scans including $100$ subjects wearing $10$ garment classes in different sizes, where we make available, scans, clothing segmentation, SMPL+G registrations, body shape under clothing, garment class and size labels.
	\item [$\bullet$] SizerNet: The first model learned from real scans to predict how clothing drapes on the body as a function of size.
	\item [$\bullet$] ParserNet: A data-driven model to map a single mesh registration into a multi-layered representation of clothing without the need for segmentation or non-linear optimization.
\end{itemize}









\begin{figure}[h]
	\centering
	\includegraphics[width=0.98\textwidth]{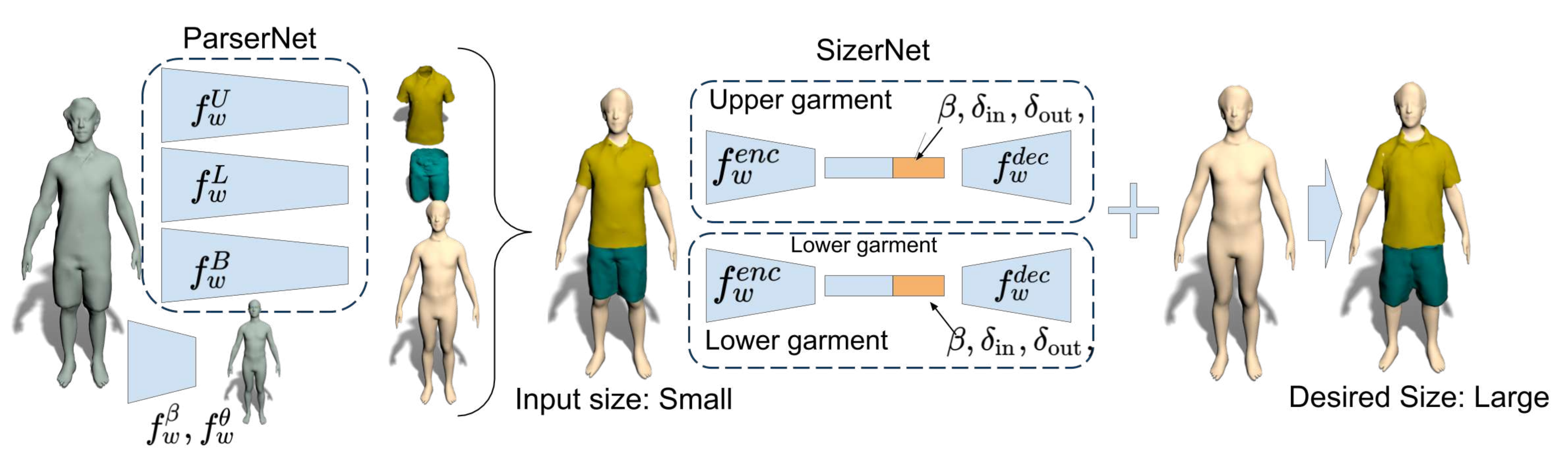}
	\caption{We propose a model to estimate and visualize the dressing effect of a garment conditioned on body shape and garment size parameters. For this we introduce \emph{ParserNet} $(\garnet^U, \garnet^L, \garnet^B)$, which takes a SMPL registered mesh $\smpl(\pose, \shape, \offsets)$ as input and predicts the SMPL parameters ($\pose, \shape$), parsed 3D garments using predefined templates $\posefun^g(\shape, \pose, \mat{0})$ and predicts body shape under clothing while preserving the personal details of the subject. We also propose \emph{SizerNet}, an encoder-decoder ($\garnet^{\mathrm{enc}}, \garnet^{\mathrm{dec}}$) based network, that resizes the garment given as input with the desired size label ($\delta_{\mathrm{in}}, \delta_{\mathrm{out}}$) and drapes it on the body shape under clothing.}
	\label{fig:overview}
\end{figure}

\section{Related Work}
\label{sec:related_work}
\noindent \textbf{Clothing modeling. }
Accurate reconstruction of 3D cloth with fine structures (e.g., wrinkles)
is essential for realism while being notoriously challenging.
Methods based on multi-view stereo can recover global shape robustly but struggle with high frequency details in non-textured regions~\cite{white2007cloth,starck2007cga,bradley2008markerless,aguiar2008siggraph,tung2009iccv,leroy2017mvdynamic}. The pioneering work of~\cite{alldieck2018video,alldieck20183DV} demonstrated
for the first time detailed body and clothing reconstruction from monocular video using a displacement from SMPL, which spearheaded recent developments~\cite{habermann2019TOG,alldieck19cvpr,alldieck2019tex2shape,saito2019pifu,habermann20deepcap,huang2020arch}. These approaches do not separate body from clothing. In~\cite{ponsmoll2017clothcap,laehner2018deepwrinkles,bhatnagar2019mgn,jiang2020bcnet}, the authors propose to reconstruct clothing as a layer separated from the body. 
These models are trained on 3D scans of real clothed people data and produce realistic models.
On the other hand, physics based simulation methods have also been used to model clothing~\cite{Wang:2010:EBW,Wang:2011:DDE,Miguel2012clothsim,DRAPE2012,stuyck2018cloth,SimulCap19,patel2020,santesteban2019virtualtryon,gundogdu19garnet}.
Despite the potential gap with real-world data, they are a great alternative to obtain clean data, free of acquisition noise and holes. However, they still require manual parameter tuning (e.g., time step for better convergence, sheer and stretch for better deformation effects, etc.), and can be slow or unstable. In~\cite{santesteban2019virtualtryon,gundogdu19garnet,DRAPE2012} a pose and shape dependent clothing model is introduced, and~\cite{patel2020,garmentdesign_Wang_SA18} also model garment style dependent clothing using a lower-dimensional representation for style and size like PCA and garment sewing parameters, however there is no direct control on the size of clothing generated for given body shape.
In~\cite{xu2020predicting}, authors model the garment fit on different body shapes from images.
Our model \emph{SizerNet} automatically outputs realistic 3D cloth models conditioned on desired features (e.g., shape, size).\\

\noindent \textbf{Shape under clothing. }
In~\cite{bualan2008naked,zhang2017detailed,yang2018analyzing}, the authors propose to estimate body shape under clothing by fitting a 3D body model to 3D reconstructions of people.
An objective function typically forces the body to be inside clothing while being close to the skin region. These methods cannot generalize well to complex or loose clothing without additional prior or supervision~\cite{chen2019arxiv}.
In~\cite{hmrKanazawa17,omran2018neural,xu2019denserac,cmr2019,SPIN:ICCV:2019,xiang2019monocular}, the authors propose learned models to estimate body shape from 2D images of clothed people, but shape accuracy is limited due to depth ambiguity.
Our model \emph{ParserNet} takes as input a 3D mesh and outputs 3D bodies under clothing with high fidelity while preserving subject identity (e.g., face details).\\

\noindent \textbf{Cloth parsing. }
The literature has proposed several methods for clothed human understanding.
In particular, efficient cloth parsing in 2D has been achieved using supervised learning and generative networks~\cite{yamaguchi2012parsing,yamaguchi2013paperdoll,yang2014cvpr,dong2019iccv,dong2020cvpr,Gong2018InstancelevelHP}.
3D clothing parsing of 3D scans has also been investigated~\cite{ponsmoll2017clothcap,bhatnagar2019mgn}. The authors propose techniques based on MRF-GrabCut~\cite{rother2004grabcut} to segment 3D clothing from 3D scans and transfer them to different subjects. However the approach requires several steps, which is not optimal for scalability.
We extend previous work with \emph{SIZER}, a fully automatic data-driven pipeline. In~\cite{bhatnagar2020ipnet}, the authors jointly predict clothing and inner body surface, with semantic correspondences to SMPL. However, it does not have semantic clothing information.\\

\noindent \textbf{3D datasets. }
To date, only a few datasets consist of 3D models of subjects with segmented clothes. 
3DPeople~\cite{pumarola20193dpeople}, Cloth3D~\cite{bertiche2019cloth3d} consists of a large dataset of synthetic 3D humans with clothing. None of the synthetic datasets contains realistic cloth deformations like the SIZER dataset.
THUman~\cite{Zheng2019DeepHuman} consists of sequences of clothed 3D humans in motion, captured with a consumer RGBD sensor (Kinectv2), and are reconstructed using volumetric SDF fusion~\cite{tao2018DoubleFusion}. However, 3D models are rather smooth compared to our 3D scans and no ground truth segmentation of clothing is provided.
Dyna and D-FAUST~\cite{pons2015dyna,dfaust:CVPR:2017} consist of high-res 3D scans of 10 humans in motion with different shape but the subjects are only wearing minimal clothing.
BUFF~\cite{zhang2017detailed} contains high-quality 3D scans of 6 subjects with and without clothing.
The dataset is primarily designed to train models to estimate body shape under clothing and doesn't contain garments segmentation.
In~\cite{bhatnagar2019mgn}, the authors create a digital wardrobe with 3D templates of garments to dress 3D bodies. In~\cite{jiang2020bcnet}, authors propose a mixture of synthetic and real data, which contains garment, body shape and pose variations. However, the fraction of real dataset ($\sim$300 scans) is fairly small. DeepFahsion3D~\cite{zhu2020deep} is a dataset of real scans of clothing containing various garment styles. None of these datasets contain garment sizing variation. Unlike our proposed \emph{SIZER} dataset, no dataset contains a large amount of pre-segmented clothing from 3D scans at different sizes, with corresponding body shapes under clothing. 




\section{Dataset}
In this paper, we address a very challenging problem of modeling garment fitting as a function of body shape and garment size. As explained in Sec.~\ref{sec:related_work}, one of the key bottlenecks that hinder progress in this direction is the lack of real-world datasets that contain calibrated and well-annotated garments in different sizes draped on real humans. To this end, we present \emph{SIZER} dataset, a dataset of over $2000$ scans containing people in diverse body shapes in various garments styles and sizes. We describe our dataset in Sec.~\ref{sec:dataset} and \ref{sec:data_define}.

\subsection{SIZER dataset: Scans}
\label{sec:dataset}
We introduce the \emph{SIZER} dataset that contains $100$ subjects, wearing the same garment in $2$ or $3$ garment sizes (S, M, L, XL). We include $10$ garment classes, namely shirt, dress-shirt, jeans, hoodie, polo t-shirt, t-shirt, shorts, vest, skirt, and coat, which amounts to roughly $200$ scans per garment class. We capture the subjects in a relaxed A-pose to avoid stretching or tension due to pose in the garments. Figure~\ref{fig:teaser} shows some examples of people wearing a fixed set of garments in different sizes. We use a Treedy's static scanner \cite{treedy} which has $130+$ cameras, and reconstruct the scans using Agisoft's Metashape software~\cite{agisoft}. Our scans are high resolution and are represented by meshes, which have different underlying graph connectivity across the dataset, and hence it is challenging to use this dataset directly in any learning framework. We preprocess our dataset, by registering them to SMPL~\cite{SMPL:2015}. We explain the structure of processed data in the following section. 

\subsection{SIZER dataset: SMPL and Garment registrations}
\label{sec:data_define}
To improve general usability of the \emph{SIZER} dataset, we provide SMPL+G registrations~\cite{lazova3dv2019,bhatnagar2019mgn} registrations. Registering our scans to SMPL, brings all our scans to correspondence, and provides more control over the data via pose and shape parameters from the underlying SMPL. We briefly describe the SMPL and SMPL+G formulations below.

SMPL represents the human body as a parametric function $M(\cdot)$, of pose $(\pose)$ and shape $(\shape)$. We add per-vertex displacements ($\offsets$) on top of SMPL to model deformations corresponding to hair, garments, etc. thus resulting in the SMPL model. SMPL applies standard skinning $\blendfun(\cdot)$  to a base template $\template$ in T-pose. Here, $\blendweights$ denotes the blend weights and $\offsetfun_p(\cdot)$ and $\offsetfun_s(\cdot)$ models pose and shape dependent deformations respectively.

\begin{equation}
\label{eq:smplpose}
\smpl(\shape,\pose,\offsets) = \blendfun(\posefun(\shape,\pose,\offsets), \jointfun(\shape), \pose, \blendweights)
\end{equation}
\begin{equation}
\label{eq:smplshape}
\posefun(\shape,\pose,\offsets) = \template + \offsetfun_s(\shape) + \offsetfun_p(\pose) + \offsets
\end{equation}

SMPL+G is a parametric formulation to represent the human body and garments as separate meshes.
To register the garments we first segment scans into garments and skin parts~\cite{bhatnagar2019mgn}. We refine the scan segmentation step used in~\cite{bhatnagar2019mgn} by fine-tuning the Human Parsing network~\cite{Gong2018InstancelevelHP} with a multi-view consistency loss. We then use the multi-mesh registration approach from \cite{bhatnagar2019mgn} to register garments to the SMPL+G model.
For each garment class, we obtain a template mesh which is defined as a subset of the SMPL template, given by $\posefun^g(\shape,\pose,\mat{0}) = \mat{I}^g\posefun(\shape,\pose,\mat{0})$, where $\mat{I}^g \in \mathbb{Z}_2^{m_g\times{n}}$
 is an indicator matrix, with $\mathbf{I}^g_{i,j}=1$ if garment $g$ vertex $i \in \{1 \hdots m_g\}$ is associated with body shape vertex $j \in \{1 \hdots n\}$. $m_g$ and $n$ denote the number of vertices in the garment template and the SMPL mesh respectively. Similarly, we define a garment function $G(\shape,\pose,\mathbf{D}^g)$ using Eq.~\eqref{eq:garment_pose}, where $\mathbf{D}^g$ are the per-vertex offsets from the template
\begin{equation}
\label{eq:garment_pose}
    G(\shape,\pose,\mathbf{D}^g) = \blendfun(\posefun^g(\shape,\pose,\offsets^g), \jointfun(\shape), \pose, \blendweights).
\end{equation}

 
For every scan in the \emph{SIZER} dataset, we will release the scan, segmented scan, and SMPL+G registrations, garment category and garment size label.

This dataset can be used in several applications like virtual try-on, character animation, learning generative models, data-driven body shape under clothing, size and(or) shape sensitive clothing model, etc. To stimulate further research in this direction, we will release the dataset,code and baseline models, which can be used as a benchmark in 3D clothing parsing and 3D garment resizing. We use this dataset to build a model for the task of garment extraction from single mesh (\emph{ParserNet}) and garment resizing (\emph{SizerNet}), which we describe in the next section.

\section{Method}
We introduce \emph{ParserNet} (Sec.~\ref{subsec:parsing_method}), the first method for extracting garments directly from SMPL registered meshes. For parsing garments, we first predict the underlying body SMPL parameters using a pose and shape prediction network (Sec.~\ref{subsec:poseshape_pred}) and use \emph{ParserNet} to extract garment layers and personal features like hair, facial features to create body shape under clothing.
Next, we present \emph{SizerNet} (Sec.~\ref{subsec:garmentresize}), an encoder-decoder based deep network for garment resizing. An overview of the method is shown in Fig.~\ref{fig:overview}.

\subsection{Pose and shape prediction network}
\label{subsec:poseshape_pred}
To estimate body shape under clothing, we first create the undressed SMPL body for a given clothed input single layer mesh $\smpl (\shape,\pose,\offsets)$, by predicting $\pose, \shape$ using $\posenet$ and $\shapenet$ respectively. We train $\posenet$ and $\shapenet$ with $L_2$ loss over parameters and per-vertex loss between predicted SMPL body and clothed input mesh, as shown in Eq.~\eqref{eq:pose_loss} and~\eqref{eq:shape_loss}. 
Since the reference body under clothing parameters $\pose,\shape$ obtained via instance specific optimization (Sec.~\ref{sec:data_define}) can be inaccurate, we add an additional per-vertex loss between our predicted SMPL body vertices $\smpl (\hat{\pose}, \hat{\shape}, \mat{0})$ and the input clothed mesh $\smpl (\shape,\pose,\offsets)$. This brings the predicted undressed body closer to the input clothed mesh. We observe more stable results training $\posenet$ and $\shapenet$ separately initially, using the reference $\shape$ and $\pose$ respectively.
Since the $\shape$ components in SMPL are normalized to have $\sigma=1$, we un-normalize them by scaling by their respective standard deviations $[\sigma_1, \sigma_2, \hdots, \sigma_{10} ]$ as given in Eq.~\eqref{eq:shape_loss}. 
 \begin{equation}
    \label{eq:pose_loss}
    \loss_\mathrm{\pose} =  \netweight_\mathrm{pose} || \hat{\pose} - \pose ||_2^2  +   \netweight_{v}  || \smpl (\shape,\hat{\pose},\mat{0}) -  \smpl (\shape,\pose,\offsets) ||
\end{equation}

 \begin{equation}
    \label{eq:shape_loss}
    \loss_{\shape} =  \netweight_\mathrm{shape} \sum^{10}_{i=1} \sigma_{i}  ( \hat{\shape}_i - \shape_i )^2 + \netweight_{v}  || \smpl  (\hat{\shape},\pose,\mat{0}) - \smpl (\shape,\pose,\offsets) ||
\end{equation}

Here, $\netweight_\mathrm{pose}$, $\netweight_\mathrm{shape}$ and $\netweight_{v}$ are weights for the loss on pose, shape and predicted SMPL surface. $(\hat{\pose},\hat{\shape})$ denote predicted parameters. The output is a \emph{smooth} (SMPL model) body shape under clothing.



\subsection{ParserNet}
\label{subsec:parsing_method}

\begin{figure}
	\centering
	\includegraphics[width=0.75\textwidth]{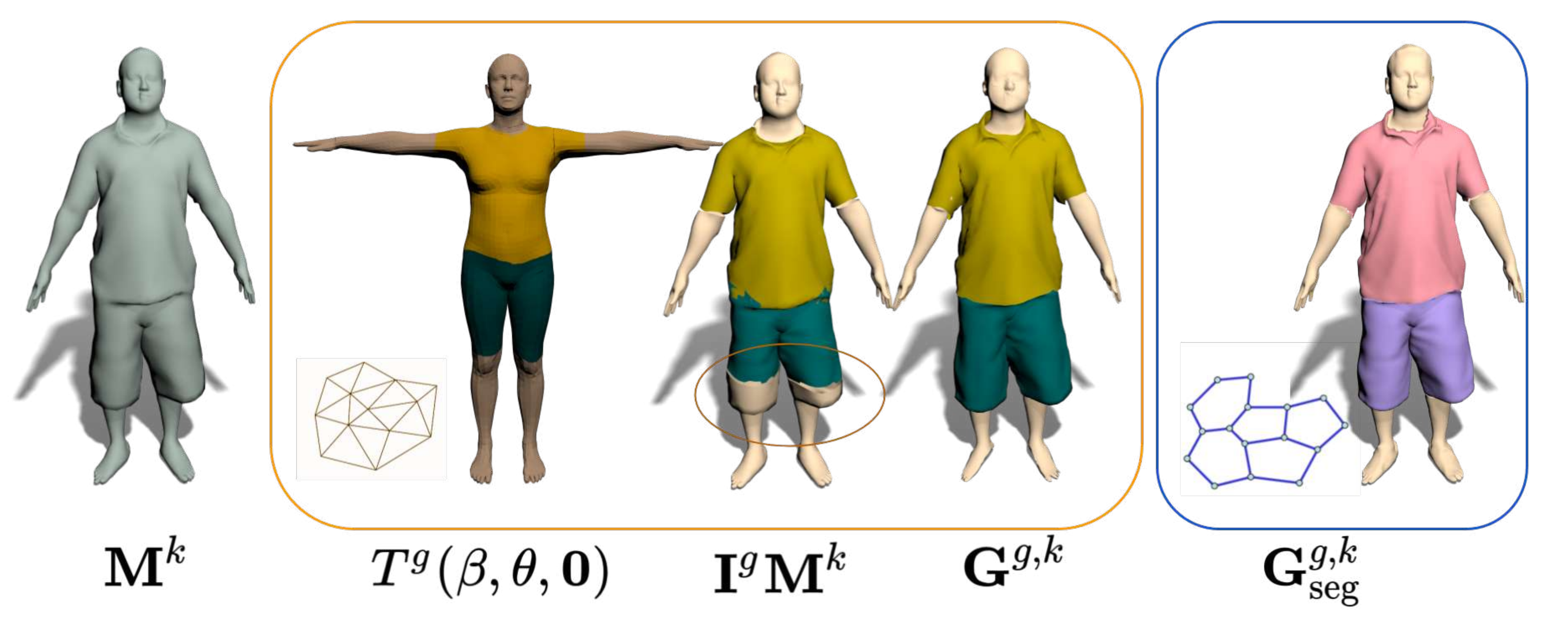}
	\caption{Left to right: Input single mesh ($\mat{M}^k$), garment template ($\posefun^g(\shape, \pose, \mat{0})=\mat{I}^g \posefun(\shape, \pose, \mat{0})$), garment mesh extracted using $\mat{G}^{g,k} = \mat{I}^g\mat{M}^k$, multi-layer meshes ($\mat{G}^{g,k}$) registered to SMPL+G, all with garment class specific edge connectivity $\mat{E}^g$, and segmented scan $\mat{G}^{g,k}_{\mathrm{seg}}$ with instance specific edge connectivity $\mat{E}^{g,k}_{\mathrm{seg}}$.}
	\label{fig:parsing_motivation}
\end{figure}

\paragraph{{\bf Parsing garments.}} Parsing garments from a single mesh ($\mat{M}$) can be done by segmenting it into separate garments for each class ($\mat{G}^{g,k}_{\mathrm{seg}}$), which leads to different underlying graph connectivity ($\mathcal{G}^{g,k}_{\mathrm{seg}} = (\mat{G}^{g,k}_{\mathrm{seg}},\mat{E}^{g,k}_{\mathrm{seg}}) $) across all the instances ($k$) of a garment class $g$, shown in Fig.~\ref{fig:parsing_motivation} (right). 
Hence, we propose to parse garments by deforming vertices of a template $\posefun^g(\shape, \pose, \mat{0})$ with fixed connectivity $\mat{E}^g$, obtaining vertices $\mat{G}^{g,k} \in \mathcal{G}^{g,k}$, where $\mathcal{G}^{g,k}= (\mat{G}^{g,k},\mat{E}^g)$, shown in Fig.~\ref{fig:parsing_motivation} (middle). 

Our key idea is to predict the deformed vertices $\mat{G}^g$ directly as a convex combination of vertices of the input mesh $\mat{M} = \smpl(\shape,\pose, \offsets)$ with a learned sparse regressor matrix $\mat{W}^g$, such that $\mat{G}^g = \mat{W}^g \mat{M}$. 
Specifically, \emph{ParserNet} predicts the sparse matrix ($\mat{W}^g$) as a function of input mesh features (vertices and normals) and a predefined per-vertex neighborhood ($\mathcal{N}_i$) for every vertex $i$ of garment class $g$. 
We will henceforth drop $(.)^{g,k}$ unless required. In this way, the output vertices $\mat{G}_i \in \mathbb{R}^3$, where $i \in \{1, \hdots, m_g\}$, are obtained as a convex combination of input mesh vertices $\mat{M}_j \in \mathbb{R}^3$ in a predefined neighborhood ($\mathcal{N}_i$).
\begin{equation}
\label{eq:fc_weight}
\mat{G}_i = \sum_{j \in \mathcal{N}_i} \mat{W}_{ij} \mat{M}_j.
\end{equation}

\paragraph{{\bf Parsing detailed body shape under clothing.}} For generating detailed body shape under clothing, we first create a \emph{smooth body mesh}, using SMPL parameters $\pose$ and $\shape$ predicted from $\posenet, \shapenet$ (Sec.~\ref{subsec:poseshape_pred}). Using the same aforementioned convex combination formulation, \emph{Body ParserNet} transfers the visible skin vertices from the input mesh to the smooth body mesh, obtaining  hair and facial features. We parse the input mesh into upper, lower garments and detailed shape under clothing using 3 sub-networks ($\garnet^U, \garnet^L, \garnet^B$) of \emph{ParserNet}, as shown in Fig.~\ref{fig:overview}.


\subsection{SizerNet}
\label{subsec:garmentresize}
We aim to edit the garment mesh based on garment size labels such as S, M, L, etc, to see the dressing effect of the garment for a new size. For this task, we propose an encoder-decoder based network, which is shown in Fig.~\ref{fig:overview} (right). The network $\garnet^{\mathrm{enc}}$, encodes the garment mesh $\mat{G}_{in}$ to a lower-dimensional latent code $\Vec{x}_{\mathrm{gar}} \in \mathbb{R}^d$, shown in Eq.~\eqref{eq:gareisze}. We append ($\shape, \delta_{\mathrm{in}}, \delta_{\mathrm{out}}$) to the latent space, where $\delta_{\mathrm{in}}, \delta_{\mathrm{out}}$ are one-hot encodings of input and desired output sizing and $\shape$ is the SMPL $\shape$ parameter for underlying body shape. 
\begin{equation}
\Vec{x}_{\mathrm{gar}} = \garnet^{\mathrm{enc}}(\mat{G}_{in}), \, \, \,  \, \, \, \garnet^{\mathrm{enc}}(.) : \mathbb{R} ^{m_g \times 3} \rightarrow \mathbb{R} ^d
   \label{eq:gareisze}
\end{equation}
The decoder network, $\garnet^{\mathrm{dec}}(.) : \mathbb{R}^{|\shape|} \times  \mathbb{R}^{d} \times \mathbb{R}^{2|\size|} \rightarrow  \mathbb{R} ^{m_g \times 3}$ predicts the displacement field $\mathbf{D}^g = \garnet^{\mathrm{dec}}(\shape, \Vec{x}_{\mathrm{gar}},\size_{\mathrm{in}} , \size_{\mathrm{out}})$ on top on template. We obtain the output garment $\mat{G}_{out}$ in the new desired size $\delta_{out}$ using Eq.~\eqref{eq:garment_pose}.



\subsection{Loss functions}
\label{subsec:trainingloss}
We train the networks, \emph{ParserNet} and \emph{SizerNet} with training losses given by Eq.~\eqref{eq:parserloss} and~\eqref{eq:sizerloss} respectively, where $\netweight_{\mathrm{3D}}, \, \netweight_{\mathrm{norm}}, \, \netweight_{\mathrm{lap}}, \, \netweight_{\mathrm{interp}}$ and $\netweight_{\mathrm{w}}$ are weights for the loss on vertices, normal, Laplacian, interpenetration and weight regularizer term respectively. We explain each of the loss terms in this section.
 
 \begin{equation}
    \label{eq:parserloss}
	\small
    \loss_{\mathrm{parser}} = \netweight_{\mathrm{3D}} \loss_{\mathrm{3D}} + \netweight_{\mathrm{norm}} \loss_{\mathrm{norm}} + \netweight_{\mathrm{lap}} \loss_{\mathrm{lap}} + \netweight_{\mathrm{interp}} \loss_{\mathrm{interp}} + \netweight_{\mathrm{w}} \loss_{\mathrm{w}}
\end{equation}

 \begin{equation}
    \label{eq:sizerloss}
	\small
    \loss_{\mathrm{sizer}} = \netweight_{\mathrm{3D}} \loss_{\mathrm{3D}} + \netweight_{\mathrm{norm}} \loss_{\mathrm{norm}} + \netweight_{\mathrm{lap}} \loss_{\mathrm{lap}} + \netweight_{\mathrm{interp}} \loss_{\mathrm{interp}} 
\end{equation}

\begin{itemize}
\item [$\bullet$] {\bf 3D vertex loss for garments. }We define $\loss_{\mathrm{3D}}$ as $L_1$ loss between predicted and ground truth vertices
\begin{equation}
    \label{eq:vertloss}
	\small
   \loss_{\mathrm{3D}}  = || \mat{G}_{\mathrm{P}} - \mat{G}_{\mathrm{GT}} ||_1.
\end{equation}


\item [$\bullet$]  {\bf 3D vertex loss for shape under clothing. }For training $\garnet^B$ (ParserNet for the body), we use the input mesh skin as supervision for predicting personal details of subject. We define a garment class specific geodesic distance weighted loss term, as shown in Eq.~\eqref{eq:bodyloss}, where $\mat{I}^s$ is the indicator matrix for skin region and $\vec{w}_{\mathrm{geo}}$ is a vector containing the $sigmoid$ of the geodesic distances from vertices to the boundary between skin and non-skin regions. The loss term is high when the prediction is far from the input mesh $\mat{M}$ for the visible skin region, and lower for the cloth region, with a smooth transition regulated by the geodesic term. Let $\mathrm{abs}_{ij}(\cdot)$ denote an element-wise absolute value operator. Then the loss is computed as

\begin{equation}
    \label{eq:bodyloss}
	\small
    \loss^{\mathrm{body}}_{\mathrm{3D}} =   \|\vec{w}^T_{\mathrm{geo}} \cdot \mathrm{abs}_{ij}(\mat{G}^s_{\mathrm{P}} - \mat{I}^s\mat{M})\|_1.
\end{equation}

    
\item [$\bullet$]  {\bf Normal Loss. }We define $\loss_{\mathrm{norm}}$ as the difference in angle between ground truth face normal ($\mat{N}^i_{\mathrm{GT}}$) and predicted face normal ($\mat{N}^i_P$).
\begin{equation}
    \label{eq:vertloss}
	\small
    \loss_{\mathrm{norm}} = \frac{1}{N_{\mathrm{faces}}} \sum^{N_{\mathrm{faces}}}_{i} (1 - (\mat{N}_{\mathrm{GT},i})^T \mat{N}_{P,i}).
\end{equation}

\item   [$\bullet$]  {\bf Laplacian smoothness term. }This enforces the Laplacian of predicted garment mesh to be close to the Laplacian of ground truth mesh. Let $\mat{L}^{g}\in \mathbb{R}^{m_g\times{m_g}}$ be the graph Laplacian of the garment mesh $\mat{G}_{\mathrm{GT}}$, and $\boldsymbol{\Delta}_{\mathrm{init}} =  \mat{L}^{g} \mat{G}_{\mathrm{GT}} \in\mathbb{R}^{m_g\times{3}}$ be the differential coordinates of the $\mat{G}_{\mathrm{GT}}$, then we compute the Laplacian smoothness term for a predicted mesh $\mat{G}_{\mathrm{P}}$ as
\begin{equation}
    \loss_{\mathrm{lap}} =   ||\boldsymbol{\Delta}_{\mathrm{init}} - \mat{L}^{g}\mat{G}_{\mathrm{P}} ||_2.
    \label{lap_loss}
\end{equation}
\GPM{What norm L1 or L2?}
\item [$\bullet$]  {\bf Interpenetration loss. }Since minimizing per-vertex loss does not guarantee that the predicted garment lies outside the body surface, we use the interpenetration loss term in Eq.~\eqref{int_loss} proposed in GarNet \cite{gundogdu19garnet}. For every vertex $\mat{G}_{\mathrm{P},j}$, we find the nearest vertex in the predicted body shape under clothing ($\mat{B}_i$) and define the body-garment correspondences as $\set{C(\mat{B}, \mat{G}_{\mathrm{P}})}$. 
Let $\mat{N}_{i}$ be the normal of the $i^{th}$ body vertex $\mat{B}_i$. If the predicted garment vertex
$\mat{G}_{\mathrm{P},j}$ penetrates the body, it is penalized with the following loss
\begin{equation}
    \loss_{\mathrm{interp}} =   \sum_{(i,j) \in \set{C(\mat{B}, \mat{G}_{\mathrm{P}})}}
    \mathbb{1}_{d(\mat{G}_{\mathrm{P},j}, \vec{G}_{\mathrm{GT},j})
     < d_{tol} } ReLU (-\mat{N}_{i} ( \mat{G}_{\mathrm{P},j} -  {\mat{B}}_{i}))/m_g,
    \label{int_loss}
\end{equation}
where notice that $\mathbb{1}_{d(\mat{G}_{\mathrm{P},j}, \vec{G}_{\mathrm{GT},j}) < d_{tol} }$ activates the loss when the distance between predicted garment mesh vertices and ground truth mesh vertices is small \emph{i.e.} $< d_{tol}$.
\item [$\bullet$]  {\bf Weight regularizer. }To preserve the fine details when parsing the input mesh, we want the weights predicted by the network to be sparse and confined in a local neighborhood. Hence, we add a regularizer which penalizes large values for $\mat{W}_{ij}$ if the distance between of $\mat{M}_j$ and the vertex $\mat{M}_k$ with largest weight $k = {\operatorname{arg\,max}}_j  \, \mat{W}_{ij}$ is large.
Let $d(\cdot,\cdot)$ dennote Euclidean distance between vertices, then the regularizer equals
\GPM{Euclidean or geodesic?} \GT{Euclidean}

\begin{equation}
\label{eq:fc_weight_local}
\loss_{w} = \sum_{i =1}^{m_g} \sum_{j \in \mathcal{N}_i} \mat{W}_{ij} d(\mat{M}_k , \mat{M}_j),  \, \, \, k = {\operatorname{arg\,max}}_j  \, \mat{W}_{ij}.
\end{equation}
\end{itemize}
\GPM{No L1 or absolute value on weights? what if negative?} \GT{weights are positive}



\subsection{Implementation Details}
\label{subsec:implement}

We implement  $\posenet$ and $\shapenet$ networks with 2 fully connected and a linear output layer. We implement \emph{ParserNet} $\garnet^U, \garnet^L, \garnet^B$  with 3 fully connected layers. We use neighborhood ($\mathcal{N}_i$) size of $|\mathcal{N}_i| = 50$, for our experiments. We first train the network for garment classes which share the same garment template and then fine-tune separately for each garment class $g$. To speed up training for \emph{ParserNet}, we train the network to predict $\mat{W}^{g} = \mat{I}^g$, where $\mat{I}^g$ is the indicator matrix for garment class $g$, explained in Sec.~\ref{sec:data_define}. This initializes the network to parse the garment by cutting out a part of the input mesh based on the constant per-garment indicator matrix, shown in Fig.~\ref{fig:parsing_motivation}. 

For \emph{SizerNet} we use $d= 30$ and we implement  $\garnet^{enc}, \garnet^{dec}$ with fully connected layers and skip connections between encoder and decoder network. We held out $40$ scans for testing in each garment class, which includes some cases with unseen subjects and some with unseen garment size for seen subjects.  For pose-shape prediction network, \emph{ParserNet} and \emph{SizerNet} we use batch-size of $8$ and learning rate of $0.0001$. 


\section{Experiments and Results}

\subsection{Results of 3D garment parsing and shape under clothing}
To validate the choice of parsing the garments using a sparse regressor matrix ($\mat{W}$), we compare the results of \emph{ParserNet} with two baseline approaches: 1) A linearized version of \emph{ParserNet} implemented with LASSO, and 
2) A naive FC network, which has the same architecture as \emph{ParserNet}. However, instead of predicting the weight matrix ($\mat{W}$), the FC network directly predicts the deformation $(\mat{D}^g)$ from the garment template ($\posefun^g(\shape, \pose, \mat{0})$) for a given input ($\mat{M}$).

\begin{figure}[t]
	\centering
	\includegraphics[width=0.99\textwidth]{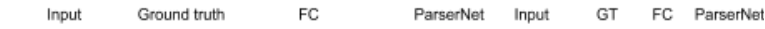}
	\includegraphics[width=0.99\textwidth]{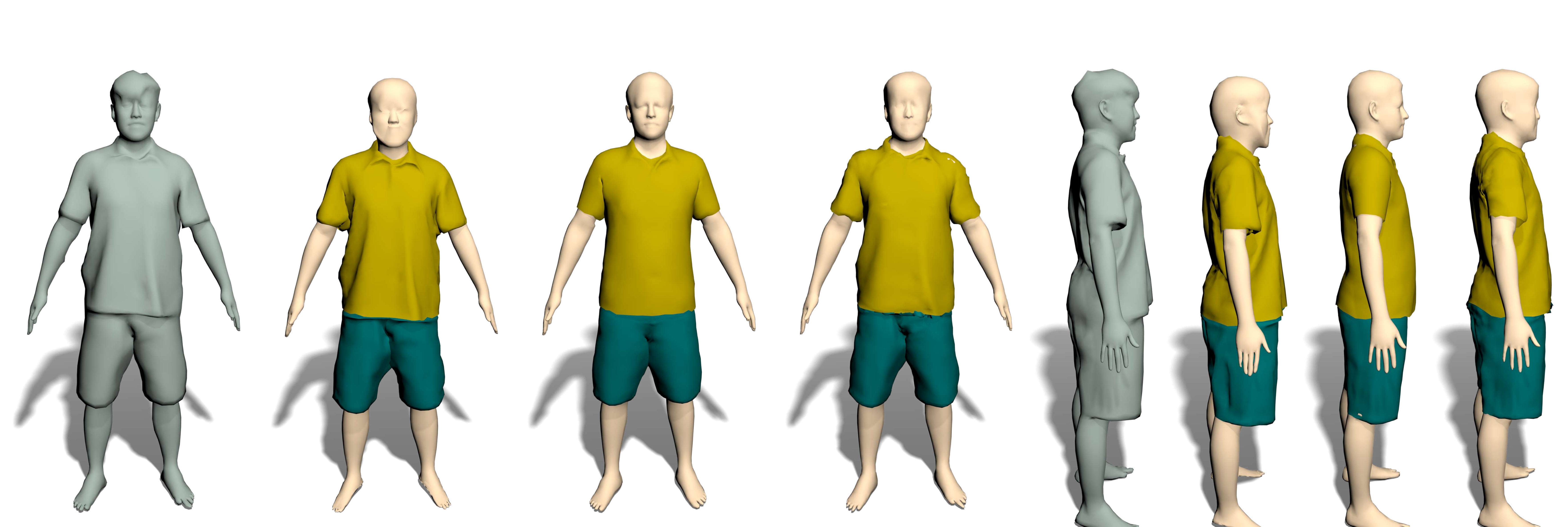}
 \includegraphics[width=0.99\textwidth]{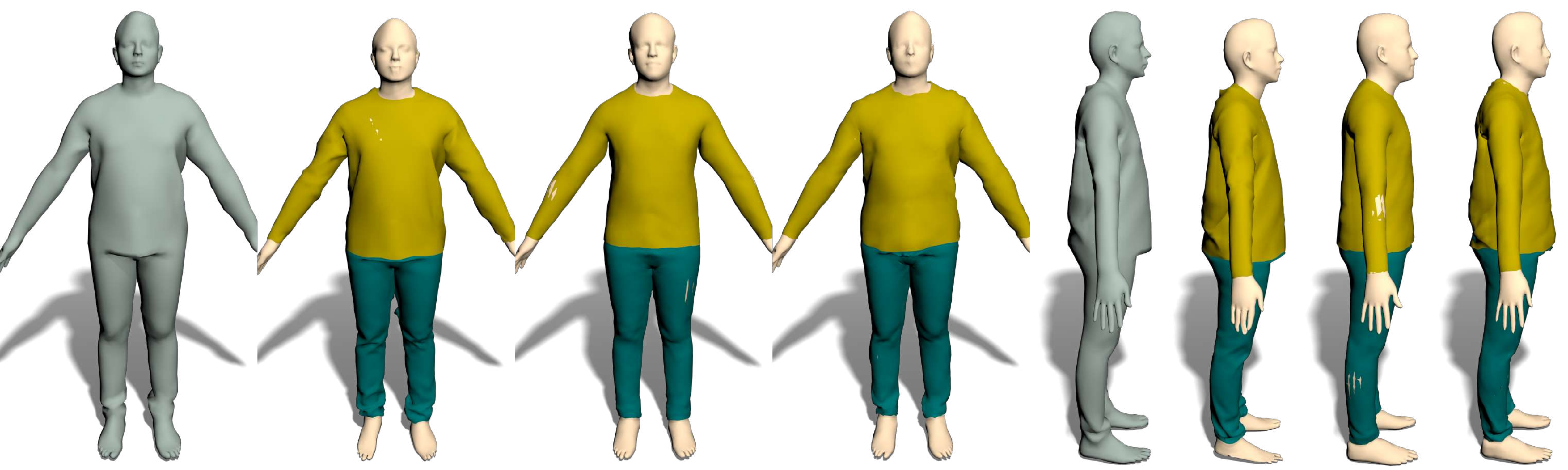}
	\caption{Comparison of \emph{ParserNet} with a FC network from front and lateral view.}
	\label{fig:parsingcomaprison}
\end{figure}

We compare the per-vertex error of \emph{ParserNet} with the aforementioned baselines in Tab.~\ref{table:extract_allmethod}. Figure~\ref{fig:parsingcomaprison} shows that \emph{ParserNet} can produce details, fine wrinkles, and large garment deformations, which is not possible with a naive FC network. This is attainable because \emph{ParserNet} reconstructs the output garment mesh as a localized sparse weighted sum of input vertex locations, and hence preserves the geometry details present in the input mesh. However, in the case of naive FC network, the predicted displacement field ($\mat{D}^g$) is smooth and does not explain large deformations. Hence, naive FC network is not able to predict loose garments and does not preserve fine details. We show results of \emph{ParserNet} for more garment classes in Fig.~\ref{fig:parsingallgarments} and add more results in the supplementary material.

\begin{table}[h]
\begin{center}

    \begin{tabular}{ |p{1.5cm}||p{1.4cm}|p{1.3cm}|p{1.5cm}||p{1.5cm}| p{1.4cm}|p{1.3cm} |p{1.5cm}|  }
 \hline
Garment & Linear Model & FC & \emph{ParserNet} & Garment & Linear Model & FC & \emph{ParserNet}  \\ 
\hline
Polo  & 32.21 & 17.25 & {\bf 14.33}  & Shorts  & 29.78 & 20.12 & {\bf 16.07}  \\
Shirt & 27.63 & 19.35 & {\bf 14.56} & Pants  & 34.82 & 18.2  & {\bf 17.24} \\
Vest  & 28.17 & 18.56 & {\bf 15.89} & Coat & 41.27  & 22.19 &  {\bf 15.34}  \\
Hoodies & 37.34 & 23.69 & {\bf 15.76} & Shorts2 & 31.38   & 23.45 & {\bf 16.23}   \\
T-Shirt & 26.94 & 15.98 & {\bf 13.77}  &   &   &  &    \\
 \hline
\end{tabular}
 \caption{Average per-vertex error $V_{\mathrm{err}}$ of proposed method for parsing garment meshes for different garment class (in mm).}
\label{table:extract_allmethod}
\end{center}

\end{table}

\begin{figure}[h]
	\centering
	\includegraphics[width=0.32\textwidth]{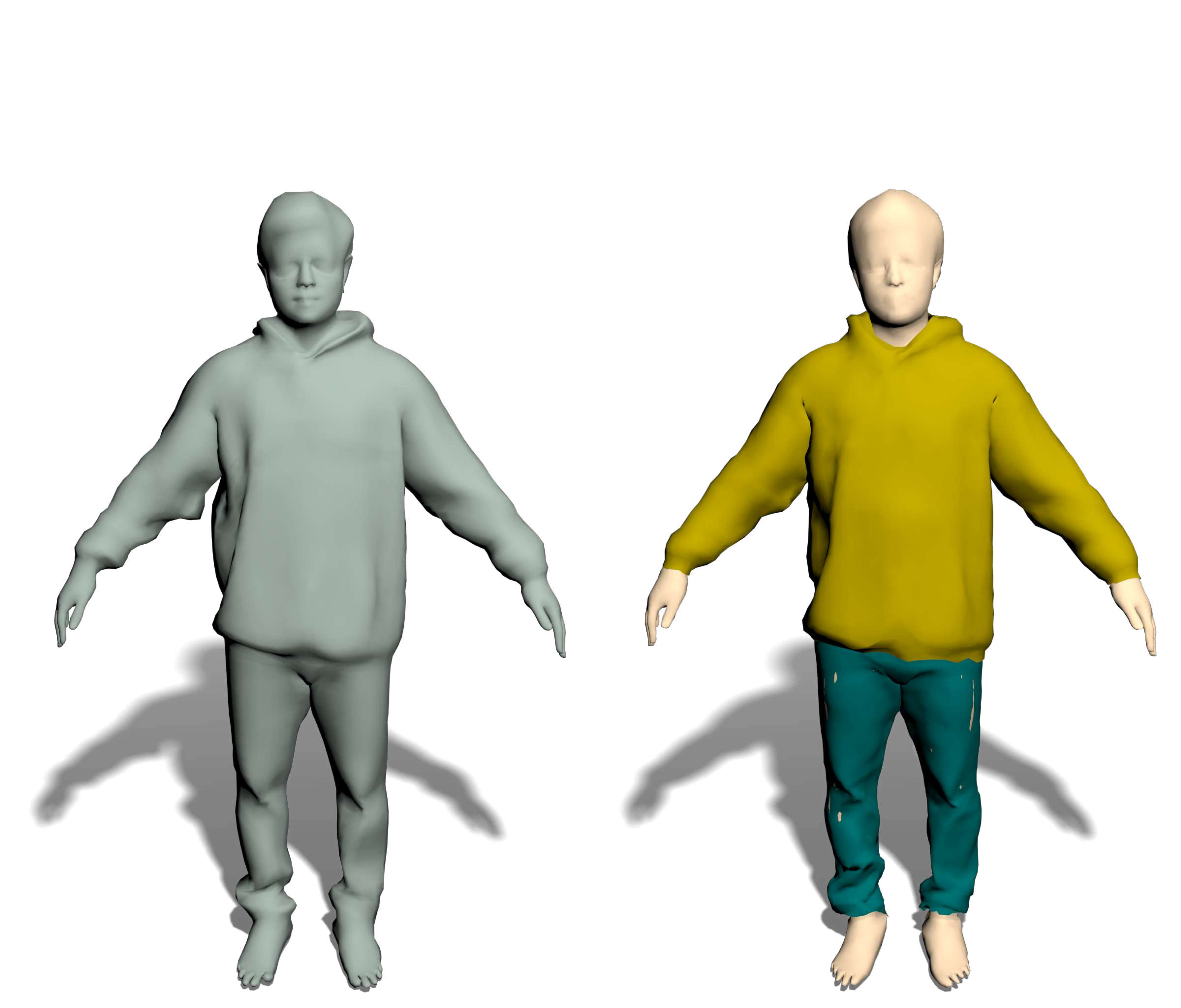}
	\includegraphics[width=0.32\textwidth]{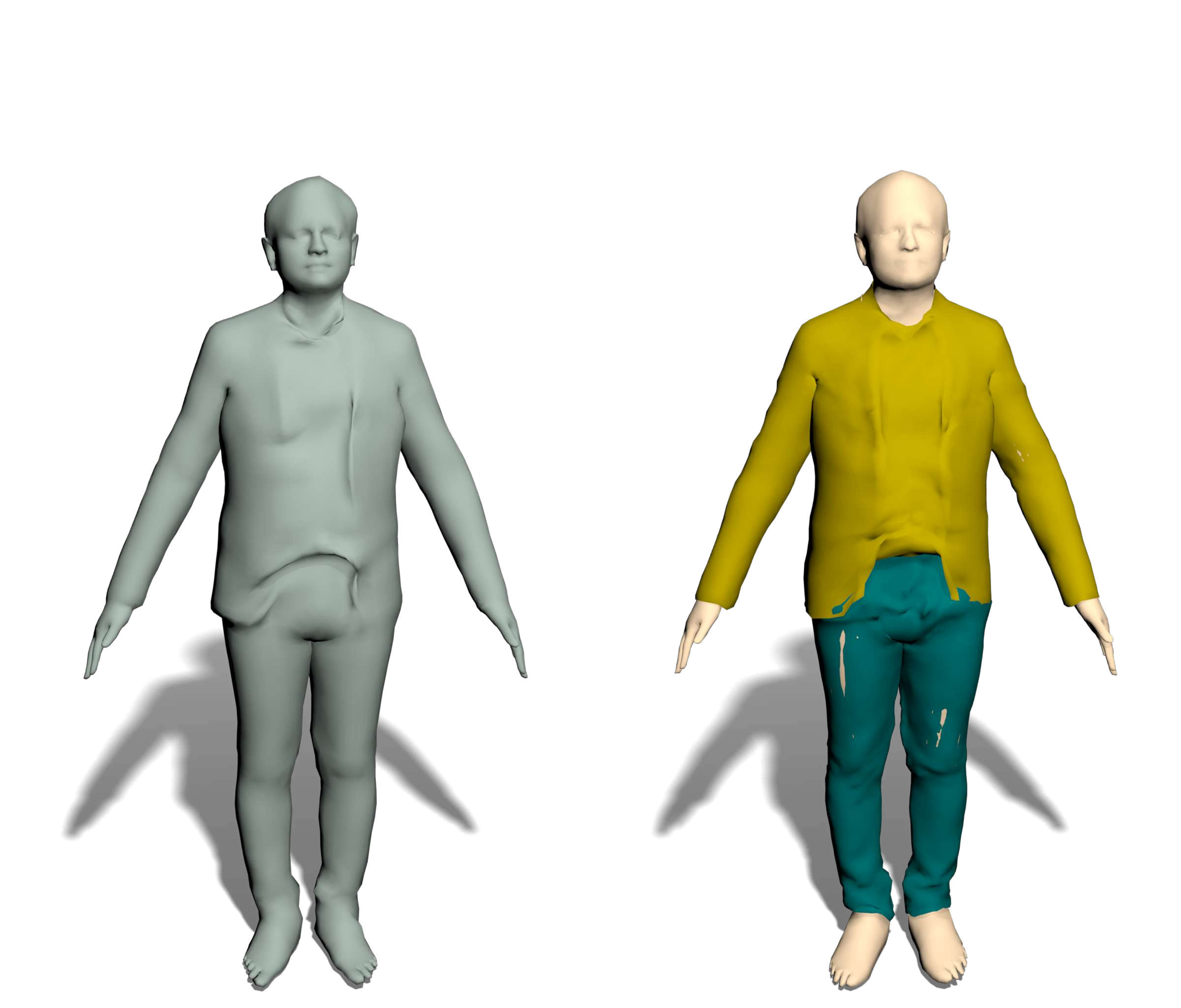}
	\includegraphics[width=0.32\textwidth]{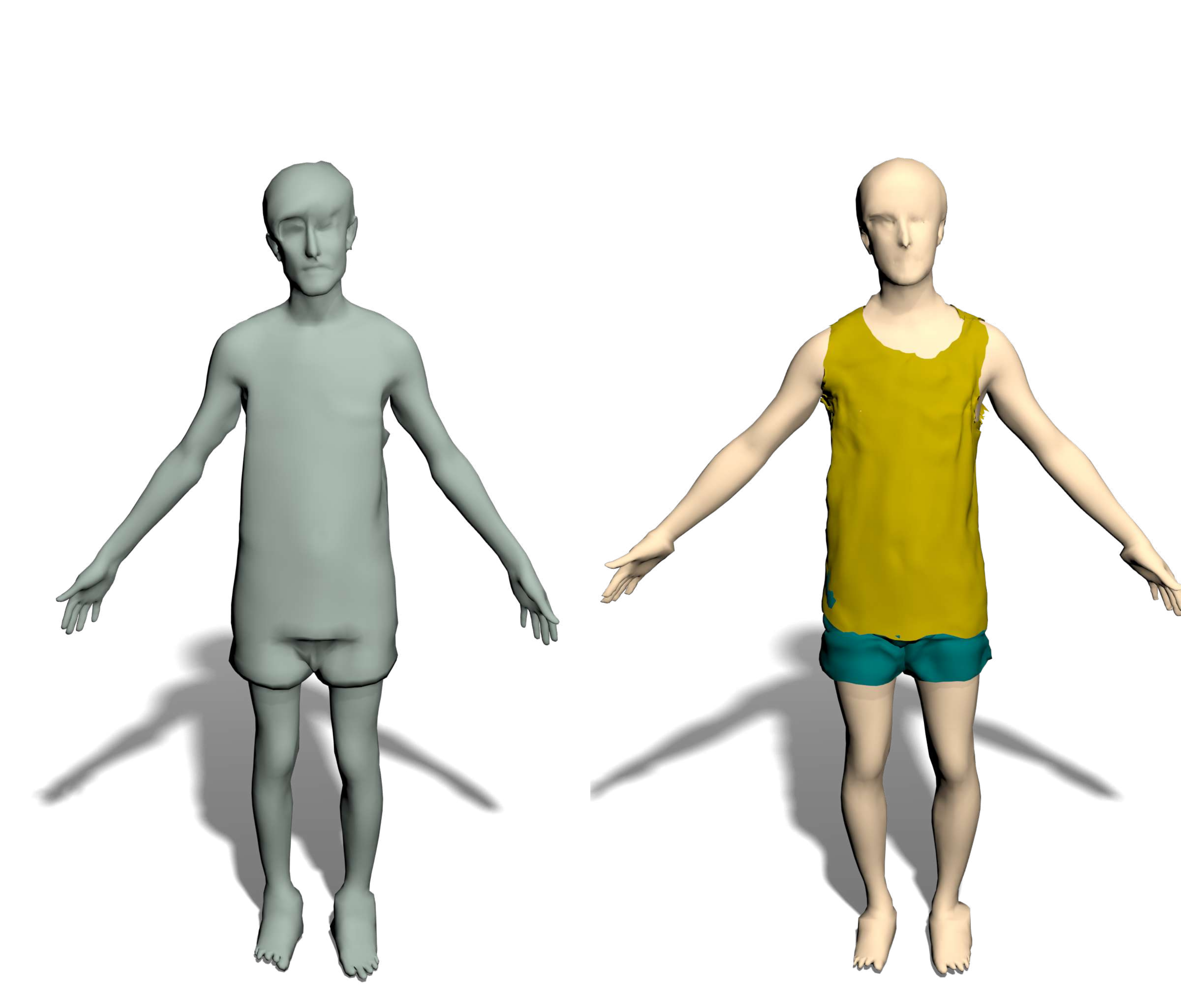}

	\includegraphics[width=0.32\textwidth]{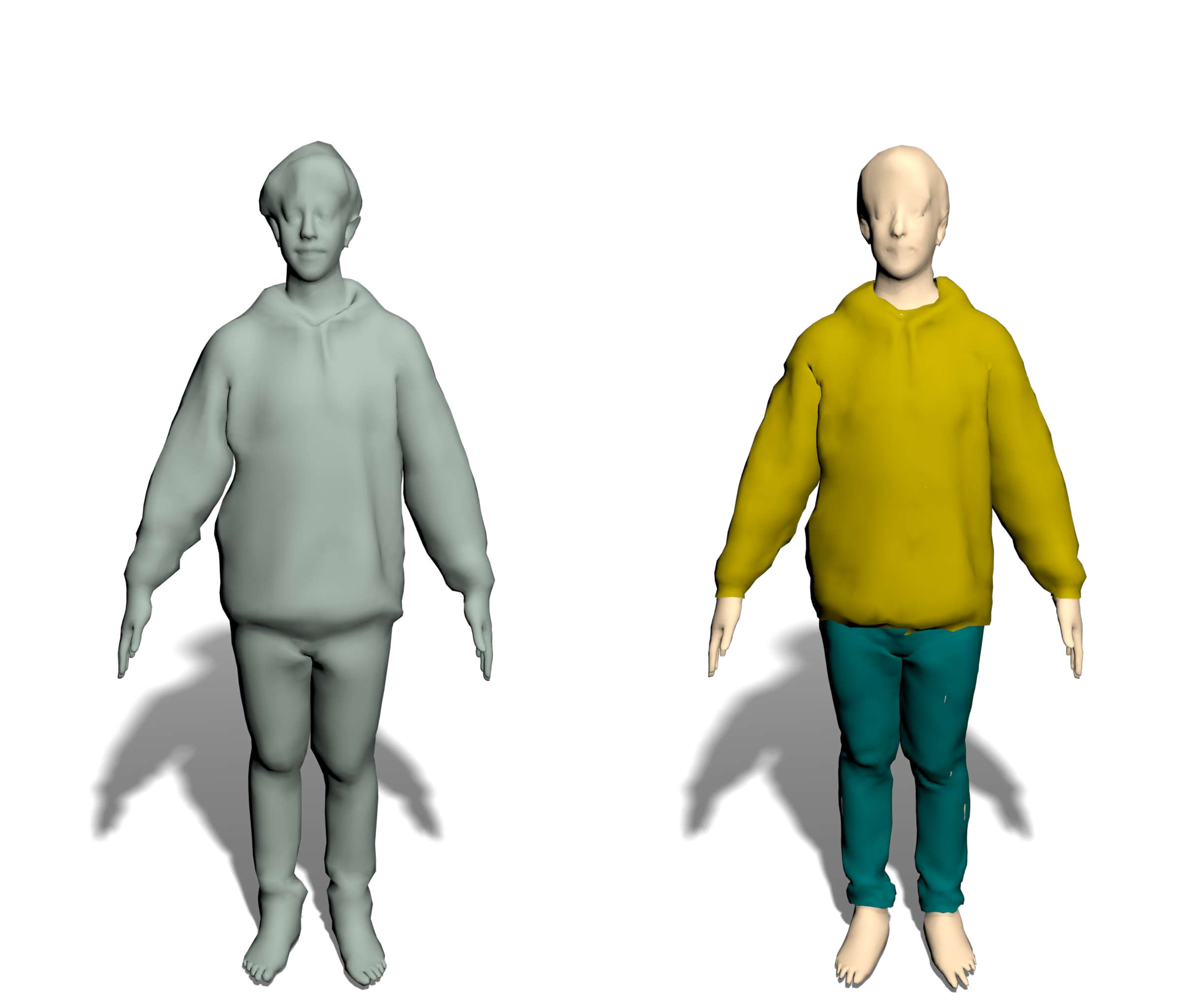}
	\includegraphics[width=0.32\textwidth]{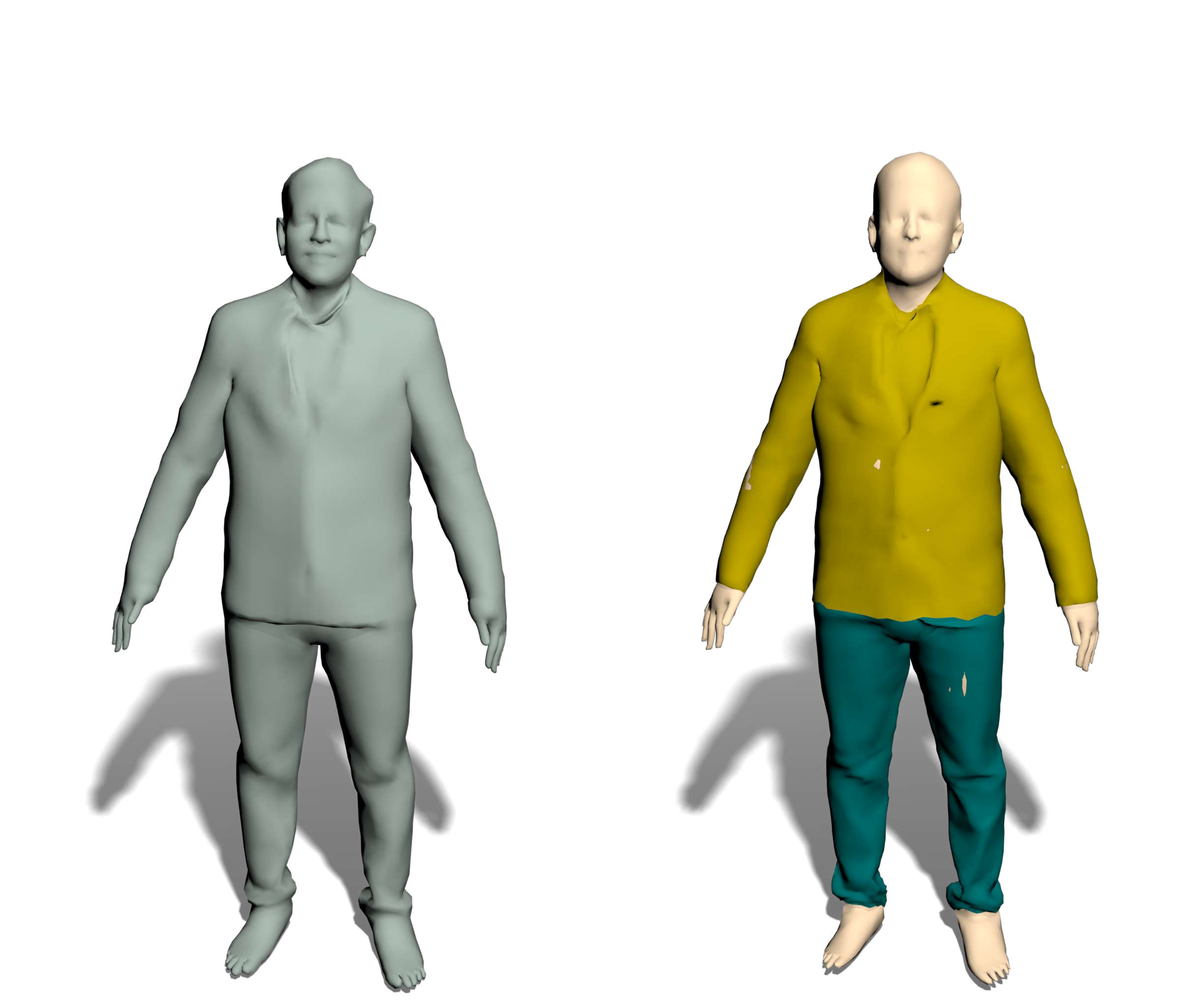}
	\includegraphics[width=0.32\textwidth]{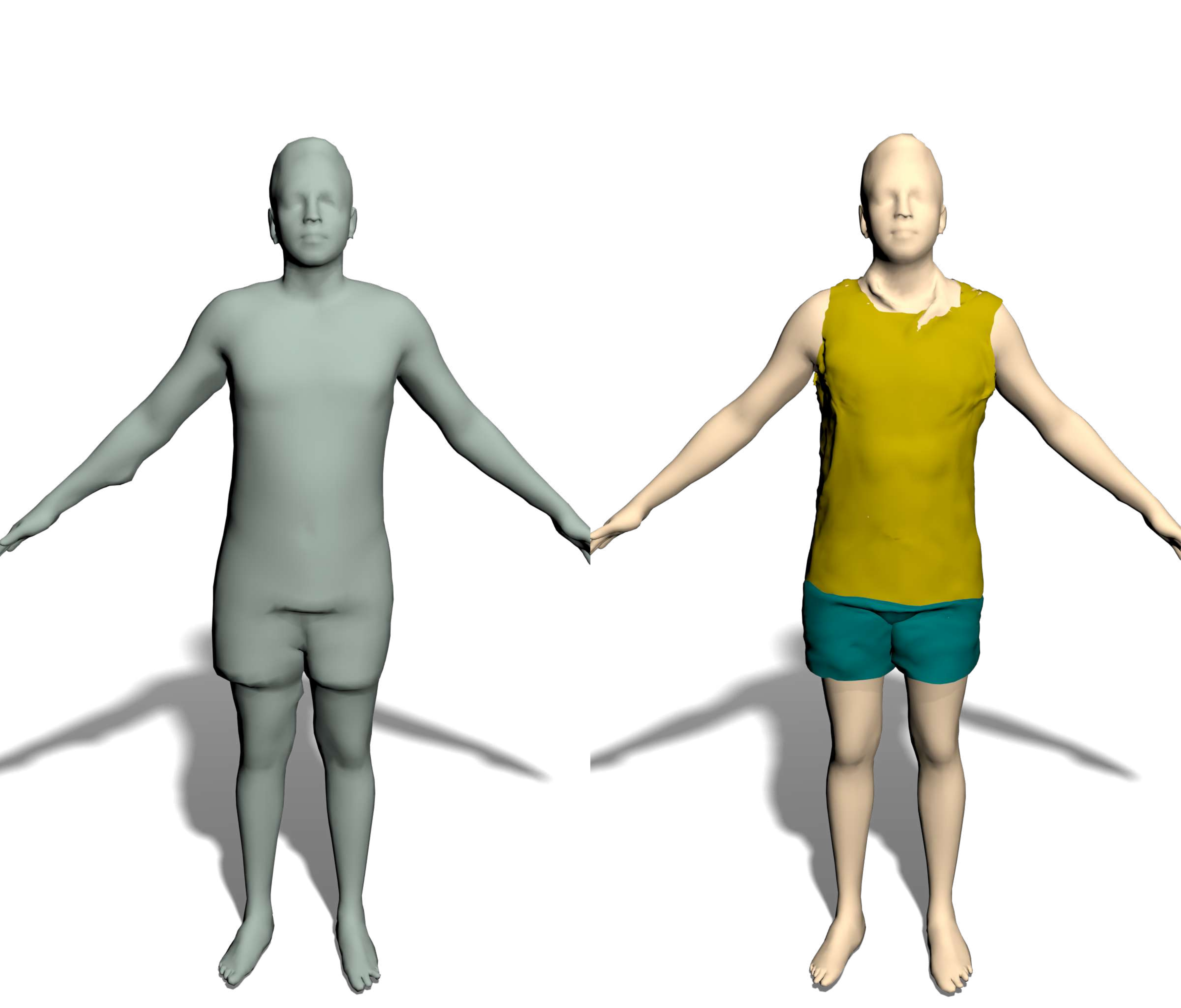}

	\caption{Input single mesh and \emph{ParserNet} results for more garments.}
	\label{fig:parsingallgarments}
\end{figure}

\subsection{Results of garment resizing}
Editing garment meshes based on garment size label is an unexplored problem and, hence there are no well defined quantitative metrics. We introduce two quantitative metrics, namely change in mesh \emph{surface area} ($A_{\mathrm{err}}$) and \emph{per-vertex error} ($V_{\mathrm{err}}$) for evaluating the resizing task. \emph{Surface area} accounts for the scale of a garment, which only changes with the garment size, and \emph{per-vertex error} accounts for details and folds created due to the underlying body shape and looseness/tightness of the garment. Moreover, subtle changes in garment shape with respect to size are difficult to evaluate. Hence, we use heat map visualizations for qualitative analysis of the results. 

Since there is no other existing work for garment resizing task to compare with, we evaluate our method against the following three baselines.
\begin{enumerate}
\item \emph{Error margin} in data: We define error margin as the change in \emph{per-vertex location} ($V_{\mathrm{err}}$) and \emph{surface area} ($A_{\mathrm{err}}$ ) between garments of two consecutive size for a subject in the dataset. Our model should ideally produce a smaller error than this margin.
\item \emph{Average prediction}: For every subject in the dataset, we create the average garment ($G_\mathrm{avg}$), by averaging over all the available sizes for a subject.
\item \emph{Linear scaling + Alignment}: We linearly scale the garment mesh, according to desired size label, and then align the garment to the underlying body.
\end{enumerate}
Table~\ref{table:size_quant} shows the errors for each experiment. \emph{SizerNet} results in lower errors, as compared to the linear scaling method, which reflects the need for modelling the non-linear relationship between garment shape, underlying body shape and garment size. We also see that network predictions yield lower error as compared to average garment prediction, which suggests that the model is learning the size variation, even though the differences in the ground truth itself are subtle. 
We present the results of \emph{SizerNet} for common garment classes in Tab.~\ref{table:size_quant}, Fig.~\ref{fig:size_resulst1},~\ref{fig:size_resulst2} and add more results in the supplementary material.
\begin{figure}[t]
	\centering
 	\includegraphics[width=0.98\textwidth]{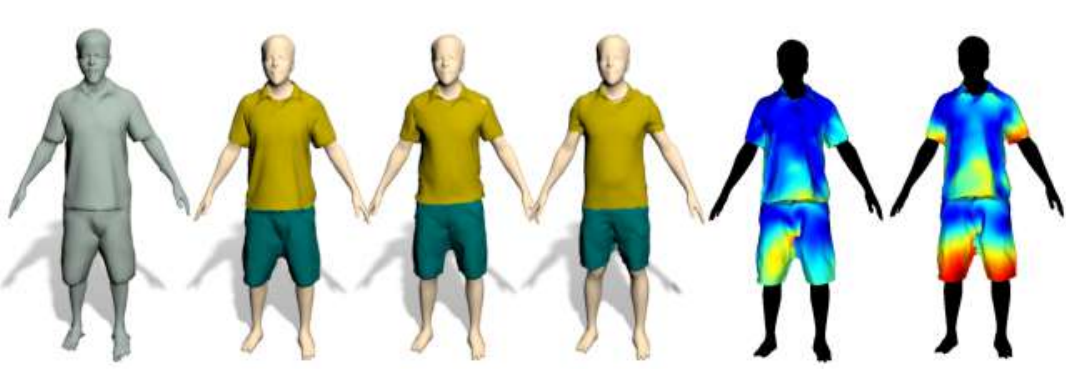}
 	\includegraphics[width=0.98\textwidth]{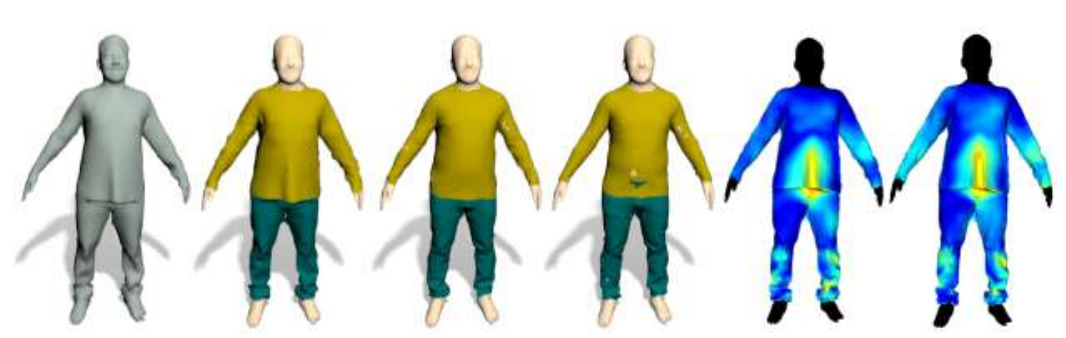}

	\caption{(a) Input single mesh. (b) Parsed multi-layer mesh from ParserNet. (c),(d) Resized garment in two subsequent smaller sizes. (e), (f) Heatmap of change in per vertex error on original parsed garment for two new sizes.}
	\label{fig:size_resulst1}
\end{figure}




\begin{table}[h]
\begin{center}

\begin{tabular}{| p{2.4cm}|p{1.1cm}|p{1.1cm} || p{1.1cm}|p{1.1cm} ||p{1.1cm}|p{1.1cm}||p{1.1cm}|p{1.1cm}|}

\hline
Garment &  \multicolumn{2}{c}{Error-margin} &    \multicolumn{2}{c}{Average-pred}  &  \multicolumn{2}{c}{Linear Scaling} &  \multicolumn{2}{c|}{Ours}  \\ 
\hline
{}   & $V_{\mathrm{err}}$  & $A_{\mathrm{err}}$     &$V_{\mathrm{err}}$  & $A_{\mathrm{err}}$     & $V_{\mathrm{err}}$  & $A_{\mathrm{err}}$    &$V_{\mathrm{err}}$  & $A_{\mathrm{err}}$ \\
Polo t-shirt   &  33.25 & 24.56    & 23.86  & 3.63 &  35.05 & 8.45   & {\bf 16.42}  & {\bf 1.79} \\
Shirt    &  36.52 &  19.57  & 21.95 & 2.76  &  34.53 & 7.01   & {\bf 15.54 } & {\bf 1.41} \\
Shorts    &  43.21&  27.21  & 24.79 & 5.41 & 35.77 & 4.99   & {\bf 16.71}  & {\bf 2.38}\\
Pants  &  30.83 & 15.15     & 21.54 & 4.73 &  38.16 & 7.13   & {\bf 19.26 } & {\bf 2.43} \\
\hline
\end{tabular}
\caption{Average per vertex error ($V_{\mathrm{err}}$ in $mm$) and surface area error($A_{\mathrm{err}}$ in $ \% $) of proposed method for garment resizing. }
\label{table:size_quant}

\end{center}
\end{table}



\begin{figure}[h]
    \centering
    \subfloat[Small(input, parsed), Medium, Large]{\includegraphics[width=0.49\textwidth]{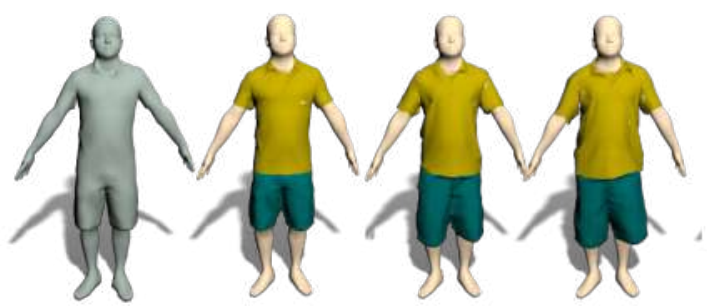}}
    \subfloat[Medium(input, parsed), Small, Large]{\includegraphics[width=0.49\textwidth]{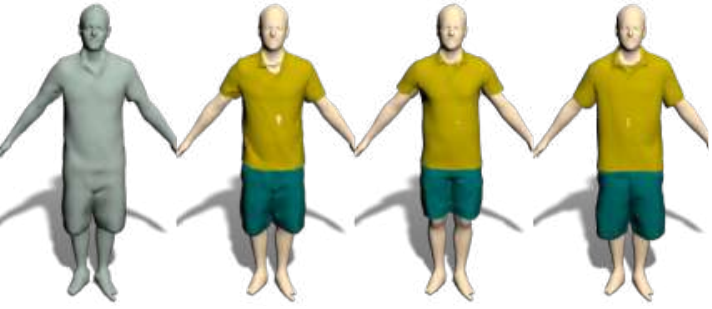}}
    
    \subfloat[Large(input, parsed), Medium, XLarge]{\includegraphics[width=0.49\textwidth]{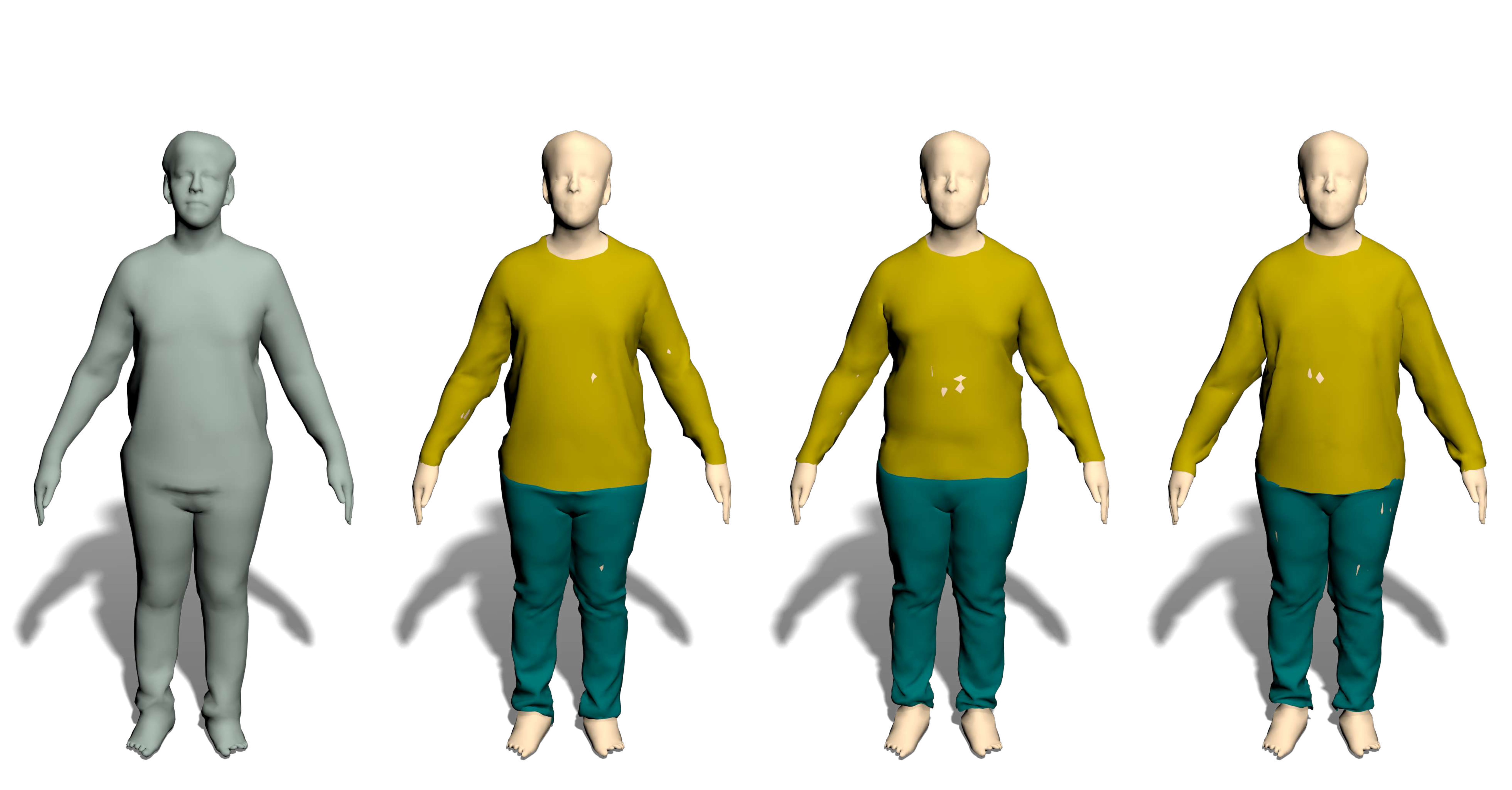}}
    \subfloat[XLarge(input,parse), Large, Medium]{\includegraphics[width=0.49\textwidth]{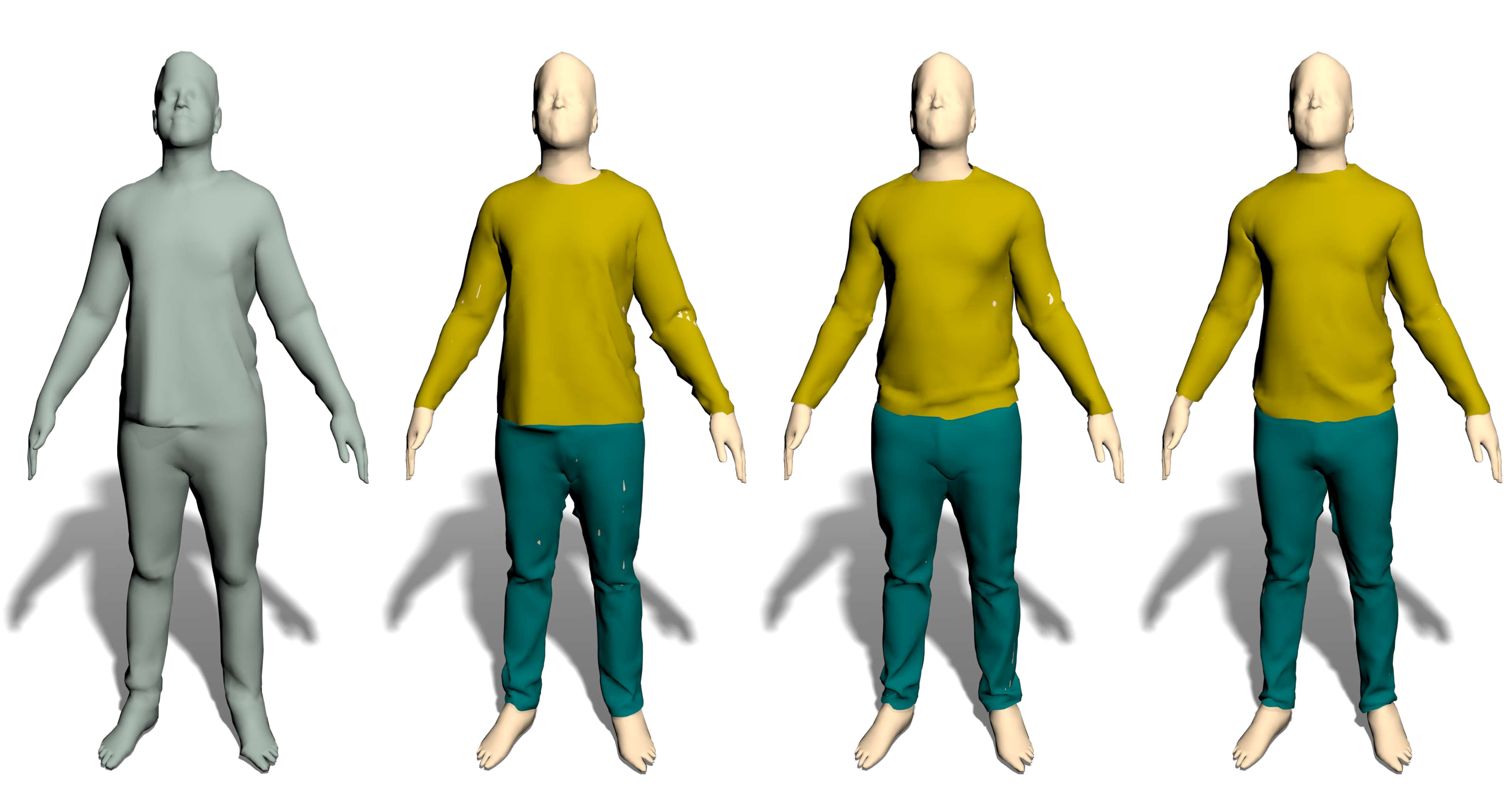}}
    \caption{Results of \emph{ParserNet} + \emph{SizerNet}, where we parse the garments from input single mesh and change the size of garment to visualise dressing effect.}%

    \label{fig:size_resulst2}%
\end{figure}

\section{Conclusion}
We introduce \emph{SIZER}, a clothing size variation dataset and model, which is the first real dataset to capture clothing size variation on different subjects. We also introduce \emph{ParserNet}: a 3D garment parsing network and \emph{SizerNet}: a size sensitive clothing model. With this method, one can change the single mesh registration to multi-layer meshes of garments and body shape under clothing, without the need for scan segmentation and can use the result for animation, virtual try-on, etc. \emph{SizerNet} can drape a person with garments in different sizes. \\
Since our dataset only consists of roughly aligned A-poses, we are limited to A-pose. We only exploit geometry information (vertices and normals) for 3D clothing parsing. In future work, we plan to use the color information in \emph{ParserNet} via texture augmentation, to improve the accuracy and generalization of the proposed method.
We will release the model, dataset, and code to stimulate research in the direction of 3D garment parsing, segmentation, resizing and predicting body shape under clothing.

\small{
\paragraph{{\bf Acknowledgements.}}This work is funded by the Deutsche Forschungsgemeinschaft (DFG, German Research Foundation) - 409792180 (Emmy Noether Programme,
project: Real Virtual Humans) and a Facebook research award. We thank Tarun, Navami, and Yash for helping us with the data capture and RVH team members~\cite{rvh_grp}, for their meticulous feedback on this manuscript.
}
\clearpage
%
%
\bibliographystyle{splncs04}
\bibliography{egbib}

\begin{thebibliography}{10}
\providecommand{\url}[1]{\texttt{#1}}
\providecommand{\urlprefix}{URL }
\providecommand{\doi}[1]{https://doi.org/#1}

\bibitem{agisoft}
Agisoft metashape, \url{https://www.agisoft.com/}

\bibitem{return_article}
The high cost of retail returns,
  \url{https://www.thebalancesmb.com/the-high-cost-of-retail-returns-2890350}

\bibitem{ihl}
Ihl group, \url{https://www.ihlservices.com/}

\bibitem{rvh_grp}
Real virtual humans, max planck institute for informatics,
  \url{https://virtualhumans.mpi-inf.mpg.de/people.html}

\bibitem{treedy}
Treedy's scanner, \url{https://www.treedys.com}

\bibitem{aguiar2008siggraph}
de~Aguiar, E., Stoll, C., Theobalt, C., Ahmed, N., Seidel, H., Thrun, S.:
  Performance capture from sparse multi-view video. {ACM} Trans. Graph.
  \textbf{27}(3),  98:1--98:10 (2008)

\bibitem{alldieck19cvpr}
Alldieck, T., Magnor, M., Bhatnagar, B.L., Theobalt, C., Pons-Moll, G.:
  Learning to reconstruct people in clothing from a single {RGB} camera. In:
  {IEEE} Conference on Computer Vision and Pattern Recognition (CVPR) (jun
  2019)

\bibitem{alldieck20183DV}
Alldieck, T., Magnor, M., Xu, W., Theobalt, C., Pons-Moll, G.: Detailed human
  avatars from monocular video. In: International Conference on 3D Vision (3DV)
  (sep 2018)

\bibitem{alldieck2018video}
Alldieck, T., Magnor, M., Xu, W., Theobalt, C., Pons-Moll, G.: Video based
  reconstruction of 3d people models. In: {IEEE} Conference on Computer Vision
  and Pattern Recognition (CVPR) (June 2018)

\bibitem{alldieck2019tex2shape}
Alldieck, T., Pons-Moll, G., Theobalt, C., Magnor, M.: Tex2shape: Detailed full
  human body geometry from a single image. In: {IEEE} International Conference
  on Computer Vision ({ICCV}). {IEEE} (oct 2019)

\bibitem{bualan2008naked}
B{\u{a}}lan, A.O., Black, M.J.: The naked truth: Estimating body shape under
  clothing. In: European Conf. on Computer Vision. pp. 15--29. Springer (2008)

\bibitem{bertiche2019cloth3d}
Bertiche, H., Madadi, M., Escalera, S.: {CLOTH3D:} clothed 3d humans. vol.
  abs/1912.02792 (2019)

\bibitem{bhatnagar2020ipnet}
Bhatnagar, B.L., Sminchisescu, C., Theobalt, C., Pons-Moll, G.: Combining
  implicit function learning and parametric models for 3d human reconstruction.
  In: European Conference on Computer Vision ({ECCV}). {Springer} (August 2020)

\bibitem{bhatnagar2019mgn}
Bhatnagar, B.L., Tiwari, G., Theobalt, C., Pons-Moll, G.: Multi-garment net:
  Learning to dress 3d people from images. In: {IEEE} International Conference
  on Computer Vision ({ICCV}). {IEEE} (oct 2019)

\bibitem{dfaust:CVPR:2017}
Bogo, F., Romero, J., Pons-Moll, G., Black, M.J.: Dynamic {FAUST}: Registering
  human bodies in motion. In: {IEEE} Conf. on Computer Vision and Pattern
  Recognition (2017)

\bibitem{bradley2008markerless}
Bradley, D., Popa, T., Sheffer, A., Heidrich, W., Boubekeur, T.: Markerless
  garment capture. In: ACM Transactions on Graphics. vol.~27, p.~99. ACM (2008)

\bibitem{chen2019arxiv}
Chen, X., Pang, A., Zhu, Y., Li, Y., Luo, X., Zhang, G., Wang, P., Zhang, Y.,
  Li, S., Yu, J.: Towards 3d human shape recovery under clothing. CoRR
  \textbf{abs/1904.02601} (2019)

\bibitem{dong2019iccv}
Dong, H., Liang, X., Wang, B., Lai, H., Zhu, J., Yin, J.: Towards multi-pose
  guided virtual try-on network. International Conference on Computer Vision
  (ICCV)  (2019)

\bibitem{dong2020cvpr}
Dong, H., Liang, X., Zhang, Y., Zhang, X., Xie, Z., Wu, B., Zhang, Z., Shen,
  X., Yin, J.: Fashion editing with adversarial parsing learning. Conference on
  Computer Vision and Pattern Recognition (CVPR)  (2020)

\bibitem{Gong2018InstancelevelHP}
Gong, K., Liang, X., Li, Y., Chen, Y., Yang, M., Lin, L.: Instance-level human
  parsing via part grouping network. In: ECCV (2018)

\bibitem{DRAPE2012}
Guan, P., Reiss, L., Hirshberg, D., Weiss, A., Black, M.J.: {DRAPE: DRessing
  Any PErson}. ACM Trans. on Graphics (Proc. SIGGRAPH)  \textbf{31}(4),
  35:1--35:10 (Jul 2012)

\bibitem{gundogdu19garnet}
Gundogdu, E., Constantin, V., Seifoddini, A., Dang, M., Salzmann, M., Fua, P.:
  Garnet: A two-stream network for fast and accurate 3d cloth draping. In:
  {IEEE} International Conference on Computer Vision ({ICCV}). {IEEE} (oct
  2019)

\bibitem{habermann2019TOG}
Habermann, M., Xu, W., , Zollhoefer, M., Pons-Moll, G., Theobalt, C.: Livecap:
  Real-time human performance capture from monocular video (oct 2019)

\bibitem{habermann20deepcap}
Habermann, M., Xu, W., , Zollhoefer, M., Pons-Moll, G., Theobalt, C.: Deepcap:
  Monocular human performance capture using weak supervision. In: {IEEE}
  Conference on Computer Vision and Pattern Recognition (CVPR). {IEEE} (jun
  2020)

\bibitem{huang2020arch}
Huang, Z., Xu, Y., Lassner, C., Li, H., Tung, T.: Arch: Animatable
  reconstruction of clothed humans. In: Proceedings of the IEEE/CVF Conference
  on Computer Vision and Pattern Recognition. pp. 3093--3102 (2020)

\bibitem{jiang2020bcnet}
Jiang, B., Zhang, J., Hong, Y., Luo, J., Liu, L., Bao, H.: Bcnet: Learning body
  and cloth shape from a single image. arXiv preprint arXiv:2004.00214  (2020)

\bibitem{hmrKanazawa17}
Kanazawa, A., Black, M.J., Jacobs, D.W., Malik, J.: End-to-end recovery of
  human shape and pose. In: Computer Vision and Pattern Regognition (CVPR)
  (2018)

\bibitem{SPIN:ICCV:2019}
Kolotouros, N., Pavlakos, G., Black, M.J., Daniilidis, K.: Learning to
  reconstruct {3D} human pose and shape via model-fitting in the loop. In:
  International Conference on Computer Vision (Oct 2019)

\bibitem{cmr2019}
Kolotouros, N., Pavlakos, G., Daniilidis, K.: Convolutional mesh regression for
  single-image human shape reconstruction. In: CVPR (2019)

\bibitem{laehner2018deepwrinkles}
Laehner, Z., Cremers, D., Tung, T.: Deepwrinkles: Accurate and realistic
  clothing modeling. In: European Conference on Computer Vision (ECCV)
  (September 2018)

\bibitem{lazova3dv2019}
Lazova, V., Insafutdinov, E., Pons-Moll, G.: 360-degree textures of people in
  clothing from a single image. In: International Conference on 3D Vision (3DV)
  (sep 2019)

\bibitem{leroy2017mvdynamic}
Leroy, V., Franco, J., Boyer, E.: Multi-view dynamic shape refinement using
  local temporal integration. In: {IEEE} International Conference on Computer
  Vision, {ICCV}. pp. 3113--3122. Venice, Italy (oct 2017)

\bibitem{SMPL:2015}
Loper, M., Mahmood, N., Romero, J., Pons-Moll, G., Black, M.J.: {SMPL}: A
  skinned multi-person linear model. ACM Trans. Graphics (Proc. SIGGRAPH Asia)
  \textbf{34}(6),  248:1--248:16 (Oct 2015)

\bibitem{ma20autoenclother}
Ma, Q., Yang, J., Ranjan, A., Pujades, S., Pons-Moll, G., Tang, S., Black, M.:
  Learning to dress 3d people in generative clothing. In: {IEEE} Conference on
  Computer Vision and Pattern Recognition (CVPR). {IEEE} (jun 2020)

\bibitem{Miguel2012clothsim}
Miguel, E., Bradley, D., Thomaszewski, B., Bickel, B., Matusik, W., Otaduy,
  M.A., Marschner, S.: Data-driven estimation of cloth simulation models.
  Comput. Graph. Forum  \textbf{31}(2),  519--528 (2012)

\bibitem{omran2018neural}
Omran, M., Lassner, C., Pons-Moll, G., Gehler, P., Schiele, B.: Neural body
  fitting: Unifying deep learning and model based human pose and shape
  estimation. In: International Conf. on 3D Vision (2018)

\bibitem{patel2020}
Patel, C., Liao, Z., Pons-Moll, G.: The virtual tailor: Predicting clothing in
  3d as a function of human pose, shape and garment style. In: {IEEE}
  Conference on Computer Vision and Pattern Recognition (CVPR). {IEEE} (Jun
  2020)

\bibitem{ponsmoll2017clothcap}
Pons-Moll, G., Pujades, S., Hu, S., Black, M.: {ClothCap}: Seamless {4D}
  clothing capture and retargeting. ACM Transactions on Graphics
  \textbf{36}(4) (2017)

\bibitem{pons2015dyna}
Pons-Moll, G., Romero, J., Mahmood, N., Black, M.J.: Dyna: a model of dynamic
  human shape in motion. ACM Transactions on Graphics  \textbf{34}, ~120 (2015)

\bibitem{pumarola20193dpeople}
Pumarola, A., Sanchez, J., Choi, G., Sanfeliu, A., Moreno-Noguer, F.:
  {3DPeople: Modeling the Geometry of Dressed Humans}. In: International
  Conference in Computer Vision (ICCV) (2019)

\bibitem{rother2004grabcut}
Rother, C., Kolmogorov, V., Blake, A.: Grabcut: Interactive foreground
  extraction using iterated graph cuts. vol.~23 (2004)

\bibitem{saito2019pifu}
Saito, S., Huang, Z., Natsume, R., Morishima, S., Kanazawa, A., Li, H.: Pifu:
  Pixel-aligned implicit function for high-resolution clothed human
  digitization. In: Proceedings of the IEEE International Conference on
  Computer Vision. pp. 2304--2314 (2019)

\bibitem{santesteban2019virtualtryon}
Santesteban, I., Otaduy, M.A., Casas, D.: {Learning-Based Animation of Clothing
  for Virtual Try-On}. Computer Graphics Forum (Proc. Eurographics)  (2019)

\bibitem{starck2007cga}
Starck, J., Hilton, A.: Surface capture for performance-based animation. {IEEE}
  Computer Graphics and Applications  \textbf{27}(3),  21--31 (2007)

\bibitem{stuyck2018cloth}
Stuyck, T.: Cloth Simulation for Computer Graphics. Synthesis Lectures on
  Visual Computing, Morgan {\&} Claypool Publishers (2018)

\bibitem{SimulCap19}
Tao, Y., Zheng, Z., Zhong, Y., Zhao, J., Quionhai, D., Pons-Moll, G., Liu, Y.:
  Simulcap : Single-view human performance capture with cloth simulation. In:
  {IEEE} Conference on Computer Vision and Pattern Recognition (CVPR) (jun
  2019)

\bibitem{tung2009iccv}
Tung, T., Nobuhara, S., Matsuyama, T.: Complete multi-view reconstruction of
  dynamic scenes from probabilistic fusion of narrow and wide baseline stereo.
  In: {IEEE} 12th International Conference on Computer Vision, {ICCV}. pp.
  1709--1716. Kyoto, Japan (Sep 2009)

\bibitem{Wang:2010:EBW}
Wang, H., Hecht, F., Ramamoorthi, R., O'Brien, J.F.: Example-based wrinkle
  synthesis for clothing animation. ACM Transactions on Graphics (Proceedings
  of {SIGGRAPH})  \textbf{29}(4),  107:1--8 (Jul 2010)

\bibitem{Wang:2011:DDE}
Wang, H., Ramamoorthi, R., O'Brien, J.F.: Data-driven elastic models for cloth:
  Modeling and measurement. ACM Transactions on Graphics (Proceedings of
  {SIGGRAPH})  \textbf{30}(4),  71:1--11 (Jul 2011)

\bibitem{garmentdesign_Wang_SA18}
Wang, T.Y., Ceylan, D., Popovic, J., Mitra, N.J.: Learning a shared shape space
  for multimodal garment design. ACM Trans. Graph.  \textbf{37}(6),  1:1--1:14
  (2018)

\bibitem{white2007cloth}
White, R., Crane, K., Forsyth, D.A.: Capturing and animating occluded cloth.
  {ACM} Trans. Graph.  \textbf{26}(3), ~34 (2007)

\bibitem{xiang2019monocular}
Xiang, D., Joo, H., Sheikh, Y.: Monocular total capture: Posing face, body, and
  hands in the wild. In: Proceedings of the IEEE Conference on Computer Vision
  and Pattern Recognition. pp. 10965--10974 (2019)

\bibitem{xu2020predicting}
Xu, H., Li, J., Lu, G., Zhang, D., Long, J.: Predicting ready-made garment
  dressing fit for individuals based on highly reliable examples. Computers \&
  Graphics  (2020)

\bibitem{xu2019denserac}
Xu, Y., Zhu, S.C., Tung, T.: Denserac: Joint 3d pose and shape estimation by
  dense render and compare. In: International Conference on Computer Vision
  (2019)

\bibitem{yamaguchi2012parsing}
Yamaguchi, K.: Parsing clothing in fashion photographs. In: Proceedings of the
  2012 IEEE Conference on Computer Vision and Pattern Recognition (CVPR). p.
  3570–3577. CVPR ’12, IEEE Computer Society, USA (2012)

\bibitem{yamaguchi2013paperdoll}
Yamaguchi, K., Kiapour, M.H., Berg, T.L.: Paper doll parsing: Retrieving
  similar styles to parse clothing items. In: {IEEE} International Conference
  on Computer Vision, {ICCV} 2013, Sydney, Australia, December 1-8, 2013. pp.
  3519--3526. {IEEE} Computer Society (2013)

\bibitem{yang2018analyzing}
Yang, J., Franco, J.S., H{\'e}troy-Wheeler, F., Wuhrer, S.: Analyzing clothing
  layer deformation statistics of 3d human motions. In: Ferrari, V., Hebert,
  M., Sminchisescu, C., Weiss, Y. (eds.) Computer Vision -- ECCV 2018. pp.
  245--261. Springer International Publishing, Cham (2018)

\bibitem{yang2014cvpr}
Yang, W., Luo, P.and~Lin, L.: Clothing co-parsing by joint image segmentation
  and labeling (2014)

\bibitem{tao2018DoubleFusion}
Yu, T., Zheng, Z., Guo, K., Zhao, J., Dai, Q., Li, H., Pons-Moll, G., Liu, Y.:
  Doublefusion: Real-time capture of human performances with inner body shapes
  from a single depth sensor. In: The IEEE International Conference on Computer
  Vision and Pattern Recognition(CVPR). IEEE (June 2018)

\bibitem{zhang2017detailed}
Zhang, C., Pujades, S., Black, M., Pons-Moll, G.: Detailed, accurate, human
  shape estimation from clothed {3D} scan sequences. In: IEEE CVPR (2017)

\bibitem{Zheng2019DeepHuman}
Zheng, Z., Yu, T., Wei, Y., Dai, Q., Liu, Y.: Deephuman: 3d human
  reconstruction from a single image. In: The IEEE International Conference on
  Computer Vision (ICCV) (October 2019)

\bibitem{zhu2020deep}
Zhu, H., Cao, Y., Jin, H., Chen, W., Du, D., Wang, Z., Cui, S., Han, X.: Deep
  fashion3d: A dataset and benchmark for 3d garment reconstruction from single
  images. arXiv preprint arXiv:2003.12753  (2020)

\end{thebibliography}
\end{document}